\documentclass[11pt]{article}

\usepackage[final]{acl}
\usepackage{graphicx}
\usepackage{amsmath}
\usepackage{booktabs}
\usepackage{amssymb}
\usepackage{tabularx}
\usepackage{microtype}
\usepackage[table]{xcolor}
\usepackage{multirow}
\usepackage{makecell}
\usepackage{rotating}
\usepackage{cuted}
\usepackage{capt-of}
\usepackage{todonotes}
\usepackage{array} 
\usepackage{placeins}
\usepackage{cuted}

\usepackage{pdflscape} 
\usepackage{caption}  
\usepackage{dblfloatfix}

\newcolumntype{C}[1]{>{\centering\arraybackslash}m{#1}}

\definecolor{ServOrange}{HTML}{E6550D}
\colorlet{rankFirst}{ServOrange!70}  
\colorlet{rankSecond}{ServOrange!55}
\colorlet{rankThird}{ServOrange!15}   

\usepackage[english,bidi=default]{babel} 
\babelfont{rm}{TeXGyreTermesX} 
\babelprovide[import]{hindi}
\babelfont[*devanagari]{rm}{Lohit Devanagari}
\babelprovide[import]{arabic}
\babelfont[*arabic]{rm}{Noto Sans Arabic}

\newcommand{\legendbox}[1]{\colorbox{#1}{\rule{0pt}{0.8ex}\hspace{0.8em}}}

\title{ServImage: An Image Generation and Editing Benchmark from Real-world Commercial Imaging Services}

\author{
\textbf{Fengxian Ji}\textsuperscript{1}\thanks{Equal contribution.},
\textbf{Jingpu Yang}\textsuperscript{2}\footnotemark[1],
\textbf{Zirui Song}\textsuperscript{1}\footnotemark[1],
\textbf{Lang Gao}\textsuperscript{1},
\textbf{Junhong Liang}\textsuperscript{1}, \\
\textbf{Zhenhao Chen}\textsuperscript{1},
\textbf{Jinghui Zhang}\textsuperscript{1},
\textbf{Xiuying Chen}\textsuperscript{1}\thanks{Corresponding author.} \\
$^{1}$MBZUAI, United Arab Emirates\\
$^{2}$Institute of Automation, Chinese Academy of Sciences, China\\
\texttt{\{fengxian.ji, zirui.song, lang.gao,junhong.liang, zhenhao.chen
\}@mbzuai.ac.ae}\\
\texttt{\{jinghui.zhang, xiuying.chen\}@mbzuai.ac.ae}\\
\texttt{\{jingpuyang290\}@gmail.com}
}

\begin{document}

\maketitle

\begin{abstract}

Recent image generation and editing models demonstrate robust adherence to instructions and high visual quality on academic benchmarks.
However, their performance on paid, real-world design projects remains uncertain. 
We introduce \textbf{ServImage}, a benchmark that explicitly correlates model outputs with economic value in commercial design projects. 
ServImage consists of 
(i) \textbf{\textit{ServImageBench}}: a dataset of 1.07k paid commercial design tasks and 2.05k designer deliverables totaling over \$295k, covering portrait, product, and digital content, along with 33k candidate images and 33k human annotations.
(ii) \textbf{\textit{ServImageScore}}: an integrated scoring system that combines three quality dimensions: baseline requirements fulfilment, visual execution quality, and commercial necessity satisfaction. These three dimensions are designed to characterize the factors that drive human payment decisions and indicate whether an image is commercially acceptable.
(iii) \textbf{\textit{ServImageModel}}: under this scoring system, we propose a payment prediction model trained on the human-annotated candidate images, achieving 82.00\% accuracy in predicting human payment decisions and producing calibrated payment probabilities.
ServImage provides a comprehensive foundation for assessing the commercial viability of image generation models and offers a scalable resource for future research on economically grounded vision systems. \url{https://github.com/FengxianJi/ServImage}

\end{abstract}

\section{Introduction}
\label{sec:intro}

\begin{figure*}[t]
\centering
\includegraphics[width=1.0\textwidth, keepaspectratio=false]{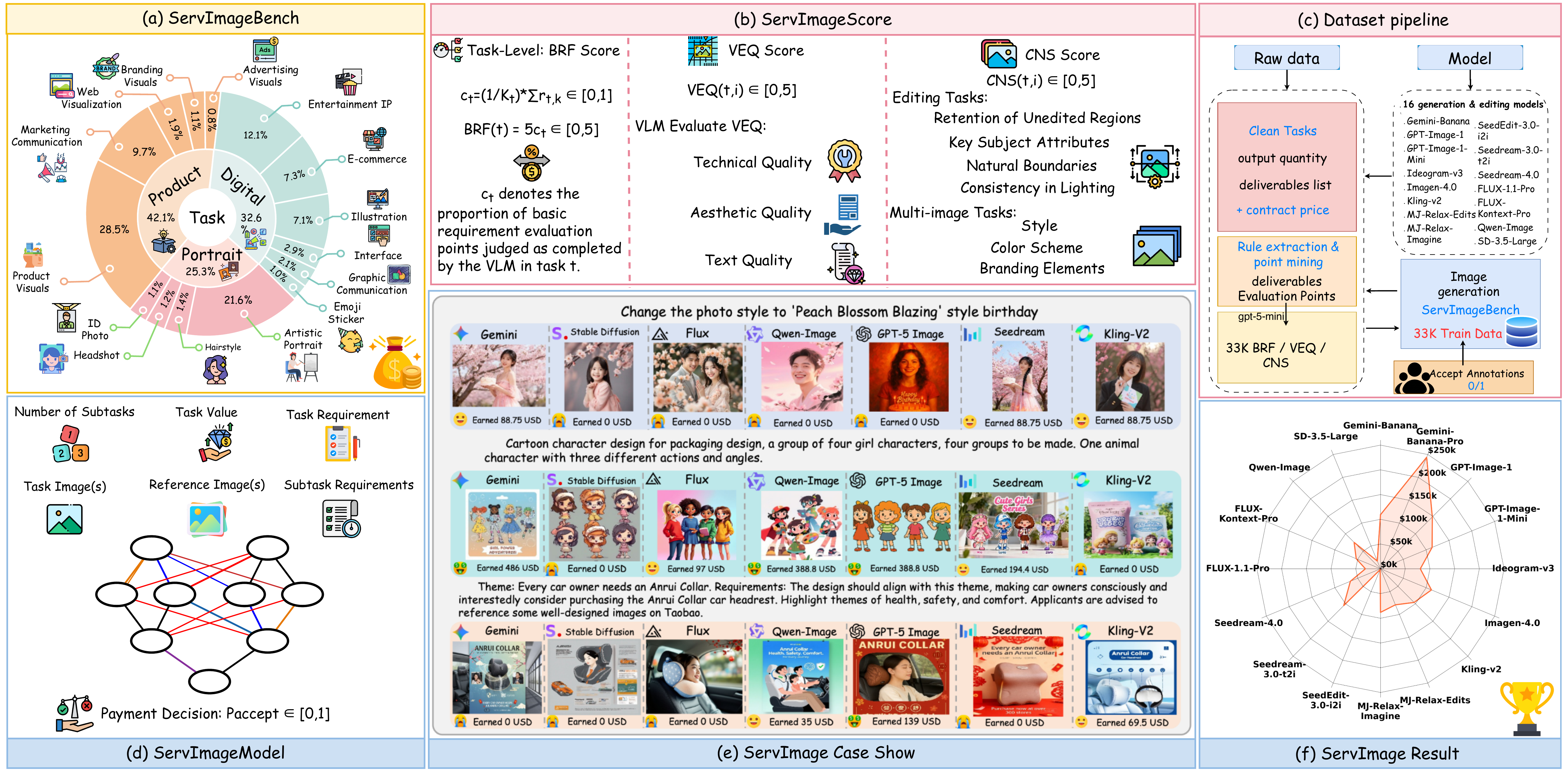}
\caption{Overview of the ServImage benchmark and evaluation framework.
(a) We collect 1{,}070 paid design tasks from online crowdsourcing platforms and group them into Portrait, Product, and Digital categories.
(b) Through the dataset pipeline with rule extraction/point mining, we obtain BRF, VEQ, and CNS scores to form ServImageScore.
(c) Sixteen image generation and editing models produce about 33k candidate images.
(d) Optionally, built on these scores, ServImageModel predicts human payment decisions.
(e) We present representative tasks and model outputs as ServImage case studies.
(f) Under the standard settlement scenario, we evaluate each model independently on the full task set and compare their commercial capability using the economic metric of total revenue.}
\label{fig:overview}
\end{figure*}

Recent years have witnessed remarkable progress in image generation and editing.
In image generation, diffusion models such as Stable Diffusion~\cite{rombach2022high} and GLIDE~\cite{nichol2021glide} have advanced high-quality text-to-image generation.
In image editing, models like InstructPix2Pix~\cite{brooks2023instructpix2pix} and MagicBrush~\cite{zhang2023magicbrush,zhang2025individuals}
have improved editing precision and controllability.
These advances have broadened the use of image generation from conventional tasks such as object removal and style transfer to more complex design tasks, including poster design, logo creation, and IP illustration~\cite{chen2024mega, wang2025omnigenbench}, which originate from real-world commercial scenarios with well-defined requirements and economic returns.
On the evaluation side, recent benchmarks emphasized instruction following and semantic controllability~\cite{ma2024i2ebench}, optical realism and reflection consistency~\cite{zeng2024dilightnet}, and physical commonsense understanding, ensuring that generated images obey plausible geometry, gravity, and material properties~\cite{farshad2023scenegenie, zhao2025envisioning, ryu2025towards, gu2025blendergym}.
However, to the best of our knowledge, no existing benchmark evaluates how well image models satisfy real-world commercial design requirements, particularly in terms of practical usability and economic acceptability.

In real design services, commercial acceptability depends on a coupled set of factors beyond semantic and perceptual correctness. 
A deliverable must first satisfy baseline business requirements, such as size, resolution, format, copyright safety, and policy compliance, before it can even be considered for delivery.
It must then achieve sufficient visual execution quality to be publishable, and, crucially, align with the client's specific commercial intent, including brand style, marketing message, and platform constraints.
Whether a client approves and pays for an image ultimately reflects these business-rule, visual, and commercial considerations jointly, which are not explicitly captured by existing benchmarks.
The fundamental question in a business context is not merely ``Is it good?'' but rather ``Is it worth paying for?''~\cite{patwardhan2025gdpval, mazeika2025remote}.

To close this gap, we introduce \textit{ServImage}, a benchmark explicitly grounded in real monetary outcomes for commercial design services. 
Concretely, \textit{ServImage} contains three components. 
\textit{\textbf{ServImageBench}} collects 1.07k paid commercial design tasks from crowdsourcing platforms, with over \$295k of contract value, together with 33k candidate images and 33k human payment decisions from 16 mainstream models, spanning portrait services, e-commerce products, and digital content, and covering both text-to-image generation and image-editing workflows.
\textbf{\textit{ServImageScore}} provides an integrated evaluation scheme that decomposes image quality into Baseline Requirements Fulfilment, Visual Execution Quality, and Commercial Necessity Satisfaction, with a unified scoring protocol that jointly captures business-rule adherence, perceptual quality, and task-specific commercial fit. 
Finally, \textbf{\textit{ServImageModel}} is a settlement model that, guided by ServImageScore and task prices, produces calibrated payment probabilities, serving as a reference method for estimating the expected revenue of different models under realistic payment rules.   
ServImage demonstrates from the perspective of commercial payment that models which appear strong under technical metrics capture only a fraction of the attainable commercial value.

In summary, our contributions include: advancing generative model evaluation from traditional technical metrics to a market-grounded perspective based on real paid tasks, human payment behavior, and price structures; uncovering the mechanisms by which commercial value is formed, showing that user preferences, task prices jointly determine economic outcomes beyond image quality alone; and providing an economic inference framework that estimates attainable revenue under realistic payment rules, revealing a substantial gap between technical performance and actual commercial value capture.
\section{Related Work}
\label{sec:Related_Work}
\paragraph{Image Generation and Editing Benchmarks.} Evaluation of generative models has progressed from traditional single-number perceptual metrics, CLIP Score, LPIPS, and PSNR/SSIM~\cite{radford2021learning, zhang2018unreasonable, wang2004image} to multi-dimensional assessment frameworks. Generation-oriented benchmarks such as GenAI-Bench and OmniGenBench foreground compositional reasoning and text rendering~\cite{li2024genai,peng2024dreambench++, zhou2025draw}, while editing-oriented ones like EDITVAL, VIEScore, and IE-Bench emphasise controllability and region consistency~\cite{basu2023editval, sun2025ie, wang2025complexbench}. Unified efforts, including I2EBench, ICE-Bench, and RISEBench, attempt to bridge this divide by jointly evaluating instruction following, visual quality, and reasoning~\cite{ma2024i2ebench, pan2025ice, xu2025manipshield}. However, a fundamental limitation persists: existing tasks lack market-validated value distributions, and current evaluations rely on conventional technical metrics that provide no insight into commercial viability.

\paragraph{VLM-based evaluation for image generation and edition.} 
To automate assessment, the VLM-as-a-Judge paradigm has become prevalent, using task-specific prompts to achieve high correlations with human preference~\cite{wu2025kris, ye2025imgedit,pu2025picabench, wang2025omnigenbench, zhang2025upme, wanglmm4lmm,yang2026frequency}.
Systems like LMM4Edit~\cite{xu2025lmm4edit,ji2025finestate} use few-shot learning for robust editing evaluation. 
While these methods align well with human perception, they don't capture the economic value of outputs. 
This gap shows a decoupling where high technical metrics don't necessarily lead to willingness to pay.
Although similar mappings from model performance to economic outcomes have been explored in other domains, such as SWE-Lancer for code generation~\cite{miserendino2025swe,li2026m3mad, yang2025agent,liang2025vision}, the image domain still lacks a verifiable, monetarily grounded framework.
Detailed comparison is in Appendix Table~\ref{tab:image_bench_comparison}.

\section{ServImage}
\label{sec:ServImage}

\subsection{Task Formulation}
\label{sec:Dataset_And_Task_Formulation}

We begin by formalizing the structure of ServImage tasks and the associated payment contracts.
Let $\mathcal{T}$ denote the set of paid design tasks in ServImage. For each task $t \in \mathcal{T}$, we denote the task context by $x_t = (\text{brief}_t, \text{refs}_t, \text{src}_t)$, where $\text{brief}_t$ is a textual description of the design goal, $\text{refs}_t$ are optional reference materials (e.g., sketches, logos, exemplar images), and $\text{src}_t$ is an optional source image when the client requests editing rather than pure generation. Each task further specifies a contract price $\text{Price}(t)$ and a required number of deliverables $Q(t)$. We define the implied per-deliverable Price as $p_{\text{img}}(t) = \text{Price}(t) / Q(t)$.

Given a task context $x_t$, a model produces a set of candidate images $\hat{Y}_t = \{\hat{\text{img}}_{t,i}\}_{i=1}^{Q(t)}$. 
Each generated image is assigned a binary validity label ($S_{t,i} \in {0,1}$) from human payment decisions, indicating whether it is accepted as a deliverable. The total value earned by the model on task $t$ is then:
\begin{equation}
    V_t(\hat{Y}_t) = \textstyle \sum_{i=1}^{Q(t)} S_{t,i} \, p_{\text{img}}(t),
\end{equation}
which measures how much the model would earn on task $t$ under the same payment contract as human designers.

\subsection{ServImageBench}  
\label{sec:ServImageBench_Introduction}
With this task formulation in place, we now describe ServImageBench, the real-world dataset from which these tasks and payment contracts are instantiated.
ServImageBench is constructed from paid design orders collected between 2018 and 2025 from two major Chinese online crowdsourcing platforms, Epwk.com and Zbj.com. Each raw posting specifies a concrete design goal, a quoted budget, and delivery requirements. In these orders, clients typically provide a textual brief and optional reference materials, expecting commercially usable visual assets tailored to a specific use case.
After cleaning, ServImage consists of 1{,}070 paid commercial design tasks and 2{,}046 deliverables, with summary statistics reported in Table~\ref{tab:servimage_stats}. 
Across the benchmark, 833 single-image tasks account for 77.9\% of the total, while the remaining 236 are multi-image tasks representing 22.1\%. We consider a task as editing if its brief includes at least one input image; otherwise, it is a pure text-to-image creation. Under this definition, 74.8\% of tasks are text-to-image creation and 25.2\% involve editing an existing image.
Representative task examples from the three categories are shown in Appendix Sec~\ref{app:task_cases}.

\begin{table}[tbp]
\centering
\footnotesize
\setlength{\tabcolsep}{4pt}
\resizebox{0.48\textwidth}{!}{%
\begin{tabular}{lrrrr}
\toprule
\textbf{Category} & \textbf{\#Tasks} & \textbf{\#Deliverables} & \textbf{Total Value} & \textbf{Avg Price} \\
\midrule
Portrait & 271 & 288 & \$15.3k & \$56.41 \\
Product & 450 & 839 & \$147.0k & \$326.65 \\
Digital & 349 & 919 & \$132.8k & \$380.40 \\
\midrule
All & 1,070 & 2,046 & \$295.0k & \$275.74 \\
\bottomrule
\end{tabular}%
}
\caption{Overall statistics of the ServImage benchmark.}
\label{tab:servimage_stats}
\end{table}

\begin{figure}[t]
    \centering   
    \includegraphics[width=\linewidth]{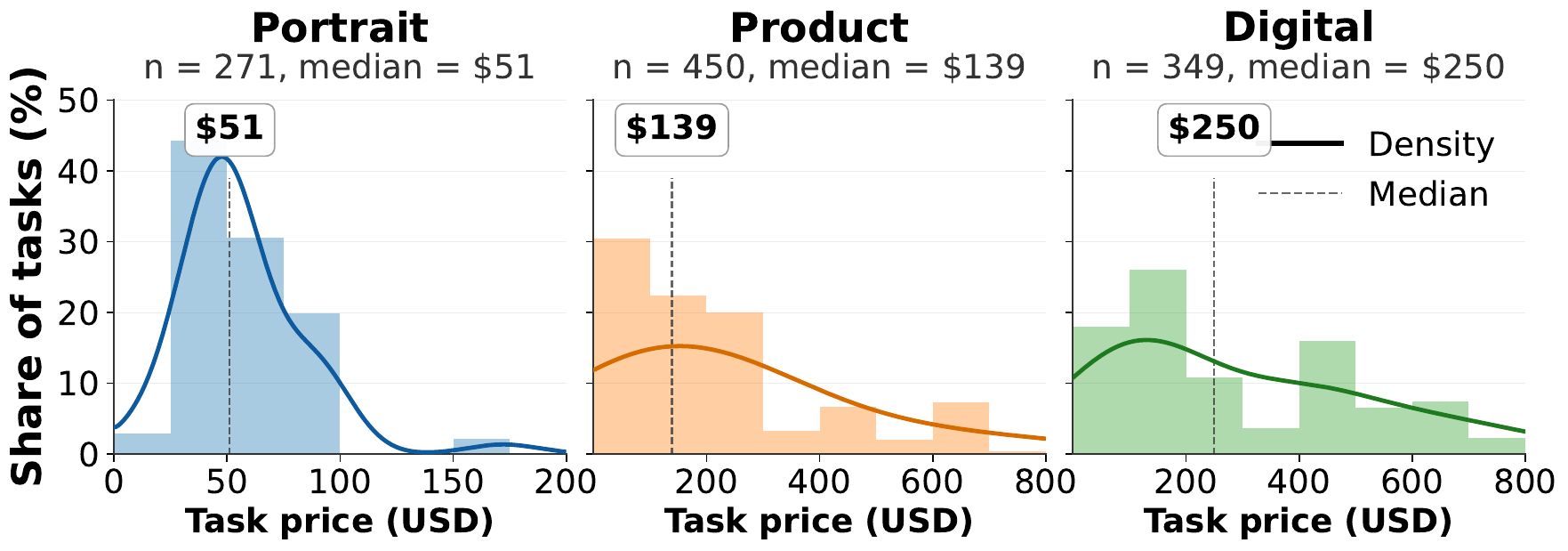}
    \caption{
    Task price distributions for Portrait, Product, and Digital categories. Dashed lines indicate median prices, showing a long-tailed pattern across categories.
    }
    \label{fig:distribution}
\end{figure}

All tasks are grouped into three client-driven commercial categories that correspond to major freelance design verticals: Portrait, Product, and Digital. 
The Portrait category covers personal imaging services such as ID photo retouching, profile pictures, and customized character portraits for social media or branding, and is dominated by single-image jobs, reflecting the one-off nature of personal imaging orders. 
Product tasks focus on e-commerce and commercial product visuals, including logo design, brand identity systems, product posters, and packaging layouts for online stores and marketing campaigns. 
Digital tasks involve broader digital content and web-oriented visual assets, such as UI mockups, web banners, illustrations, IP characters, and various online media creatives for websites, apps, or social platforms. 
Table~\ref{tab:servimage_stats} and Figure~\ref{fig:distribution} summarize the number of tasks and deliverables, as well as the total contract value across these categories.

\subsection{ServImageScore}
\label{sec:ServImageScore}

Existing benchmarks for image generation models rarely measure their economic value: their metrics mainly focus on instruction following and visual fidelity, which are not directly applicable to commercial design services. 
Starting from how real clients judge deliverables, we define three practical aspects and bundle them into a unified label tuple, \textbf{ServImageScore}, which represents commercial acceptability through BRF, VEQ, and CNS together with the deliverable-level binary payment label, rather than through a fixed weighted aggregation.
For each task $t$ with associated images $i$, ServImageScore consists of a baseline requirement fulfillment score at task-level $\text{BRF}(t) \in [0,5]$ shared by all images in $t$, and two image-level scores $\text{VEQ}(t,i), \text{CNS}(t,i) \in [0,5]$ defined for each image $(t,i)$, together with a binary payment decision $\text{Accept}(t,i)\in\{0,1\}$ indicating whether the deliverable is labeled as approved and paid under the original task contract by human annotators.

\paragraph{Baseline Requirements Fulfilment, BRF.}
BRF measures whether a candidate satisfies the explicit requirements in the client brief, such as logo placement and background colour.
We first run a rule-extraction pipeline that decomposes each brief into binary evaluation points, with the VLM predicting completion (0/1) for each by comparing the image against the brief and references.  
For multi-image tasks, an evaluation point is satisfied if at least one image meets it. 
We then compute the completion rate
$c_t = \frac{1}{K_t} \sum_{k=1}^{K_t} r_{t,k} \in [0,1]$, which represents the fraction of requirements satisfied for task $t$.
To keep BRF on the same $[0,5]$ scale as the other ServImageScore components, we linearly rescale this rate and define the task-level score as $\text{BRF}(t) = 5 c_t \in [0,5]$, and this value is shared by all images $i$ in task $t$. 
For notational uniformity at the image level, we define $\mathrm{BRF}(t,i):=\mathrm{BRF}(t)$ for all images $i$ in task $t$.

\paragraph{Visual Execution Quality, VEQ.}
VEQ assesses the overall visual quality of an image, independent of whether it follows the brief. 
For each image $i$ in task $t$, the VLM is instructed to consider three aspects: (i) technical quality (clarity, artefacts, realism), (ii) aesthetic quality (composition, colour, lighting, harmony), and (iii) text quality when textual elements are present (readability and typography).
Based on this rubric, the VLM directly outputs a single visual-quality score $\text{VEQ}(t,i) \in [0,5]$ for image $(t,i)$. 
When an image contains no textual elements, the text-quality aspect is marked as N/A and excluded from aggregation, and $\text{VEQ}(t,i)$ is computed from the remaining aspects but still lies on the same $[0,5]$ scale.

\paragraph{Commercial Necessity Satisfaction, CNS.}
CNS captures consistency in settings where relationships between images matter, namely, image editing and multi-image tasks. For editing tasks, $\text{CNS}(t, i)$ assesses the preservation of non-edited regions, key subject attributes, natural boundaries, and coherent lighting and perspective. For multi-image tasks, we compute a set-level score $\text{CNS}(t)$ for cross-image consistency in style, colour palette, and brand elements, and assign $\text{CNS}(t, i):= \text{CNS}(t)$ to all images in the task; when a task does not involve editing or multiple images, $\text{CNS}(t, i)$ is marked as N/A and excluded from score aggregation.

\subsection{Data Annotation}
\label{sec:data_annotation}

To operationalise the ServImageScore framework and obtain ground-truth labelsfor later modelling Sec.~\ref{sec:ServImageModel}, we annotate model-generated candidates across all tasks in ServImageBench. 
Concretely, we run the 16 image generation and editing models introduced in Sec.~\ref{sec:Experiments_and_Results}, denoted by $\mathcal{M}$, on all paid tasks.
For each task $t \in \mathcal{T}$ and model $m \in \mathcal{M}$, we generate $Q(t)$ candidate deliverables $\{\hat{\text{img}}_{t,i,m}\}_{i=1}^{Q(t)}$ using the task brief, reference materials, and source image if provided.
Collectively, these outputs form an extended candidate set $\mathcal{D}_{33\text{K}}$ of approximately $33\text{k}$ images.

\begin{figure}[t]
  \centering
  \includegraphics[width=0.9\linewidth]{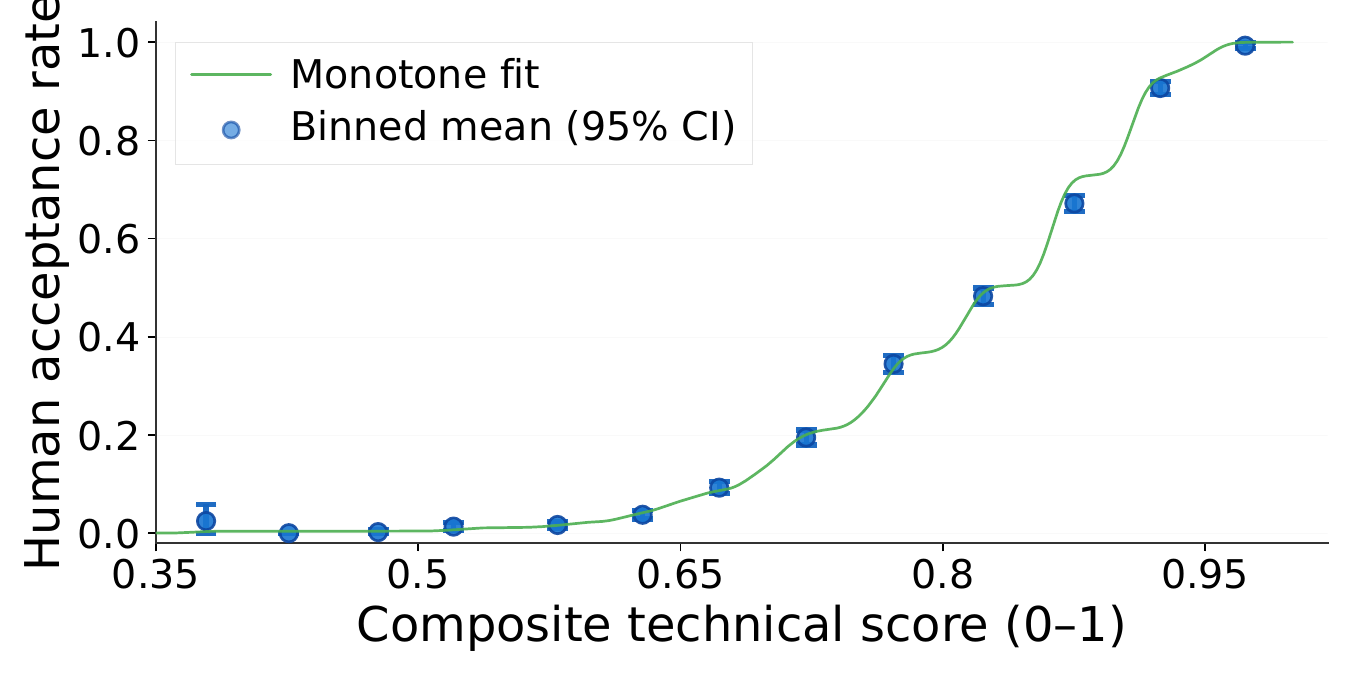}
  \caption{  Composite scores from BRF, VEQ, and CNS correlate with acceptance rates on ServImage-33K, showing that $\textbf{s}_{t,i}$ aids payment prediction. Data splits are at the task level to prevent leakage across deliverables from the same order.}
  \label{fig:score_validity}
\end{figure}

Each candidate $(t,i,m) \in \mathcal{D}_{33\text{K}}$ inherits the original task context and is annotated with the ServImageScore tuple defined in Sec.~\ref{sec:ServImageScore}.
That is, we obtain a task-level BRF score $\text{BRF}(t)$ and image-level scores $\text{VEQ}(t,i)$ and $\text{CNS}(t,i)$, all in $[0,5]$, together with a binary payment label $y_{t,i} \in \{0,1\}$, where $y_{t,i}=1$ indicates that the candidate would be approved and paid under the original task contract, and $y_{t,i}=0$ otherwise.
In all experiments, we treat $y_{t,i}$ as the ground-truth settlement outcome; ServImageModel is used only as an auxiliary predictor for expected-revenue estimation.
The triple $(\text{BRF}(t, i), \text{VEQ}(t,i), \text{CNS}(t,i))$ serves as the ground-truth concept scores for Sec.~\ref{sec:ServImageModel}, while $y_{t,i}$ is the ground-truth payment decision.
As shown in Fig.~\ref{fig:score_validity}, a composite score derived from the annotated concepts $\text{s}_{t,i}$ is monotonically correlated with empirical acceptance rates, supporting these dimensions as a meaningful concept space for payment prediction.

The BRF, VEQ, and CNS scores are obtained automatically through a unified VLM-as-a-judge pipeline described in Sec.~\ref{sec:Experiments_and_Results}, whereas the payment labels $y_{t,i}$ are assigned by trained human annotators following platform-style guidelines.
The full prompts are provided in Appendix~\ref{app:prompting}.
We adopt a double-annotation-with-adjudication protocol, following prior work on expert-labelled datasets~\citep{jin2019pubmedqa}. Each deliverable is independently labelled by two annotators. 
Across 2,000 randomly sampled doubly annotated instances, we observe an inter-annotator agreement of 67.92\%. Any remaining disagreements are resolved by a third annotator, who adjudicates the final label.
\section{ServImageModel}
\label{sec:ServImageModel}

\begin{figure*}[t]
\centering
\includegraphics[width=1\textwidth, keepaspectratio=false]{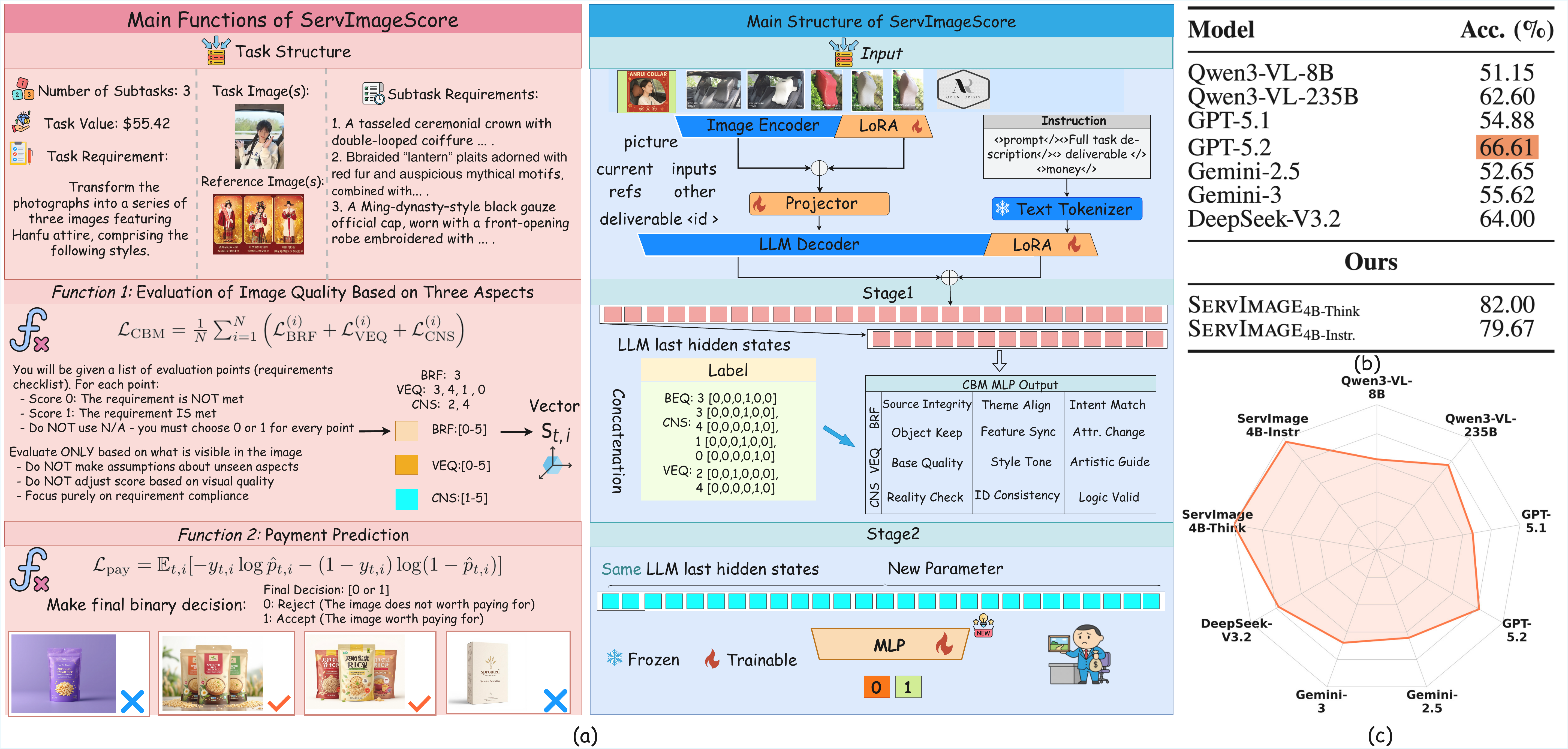}
\caption{Overview of ServImageModel: (a) Two-stage ServImageModel architecture; (b) Accuracy comparison of base models and our variants. 
(c) Radar chart of Accuracy Comparison Result.
}
\label{fig:workflow}
\end{figure*}

With the annotated candidate dataset in place, we now turn to modelling human payment decisions.
Our objective is to train ServImageModel, a predictor of whether each candidate image $\hat{\mathrm{img}}_{t,i}$ in task $t$ would be accepted and paid according to human payment labels $y_{t,i}$.
Concretely, ServImageModel is a two-stage neural network consisting of
(i) a concept predictor and
(ii) a payment head $f_{\theta}$.
Given the task context $x_t$ and a candidate image $\hat{\mathrm{img}}_{t,i}$, the model first predicts the three ServImageScore dimensions as intermediate concepts, and then uses these predicted concepts to estimate the final acceptance probability.

\paragraph{(1) Concept-bottleneck prediction from the three quality dimensions.}
In our case, the three ServImageScore dimensions act as these explicit intermediate concepts.
For each candidate $(t,i)$, we denote the annotated concept vector as $\text{s}_{t,i}
= (\mathrm{BRF}(t,i),\, \mathrm{VEQ}(t,i),\, \mathrm{CNS}(t,i))$.
During training, the model predicts a corresponding concept vector $\hat{\text{s}}_{t,i}$, and matches it to the annotated scores using regression losses
$\mathcal{L}_{\text{BRF}}^{(n)}$,
$\mathcal{L}_{\text{VEQ}}^{(n)}$, and
$\mathcal{L}_{\text{CNS}}^{(n)}$.
The concept loss is as:
\begin{equation}
\textstyle
\resizebox{0.895\linewidth}{!}{$
\mathcal{L}_{\text{CBM}}
  = \frac{1}{N} \sum_{n=1}^{N}
      \left(
        \mathcal{L}_{\text{BRF}}^{(n)}
      + \mathcal{L}_{\text{VEQ}}^{(n)}
      + \mathcal{L}_{\text{CNS}}^{(n)}
      \right).
  \label{eq:cbm_loss}
$}
\end{equation}

\paragraph{(2) Payment prediction with a task-aware head.}
On top of the learned concepts, we train a payment predictor $f_\theta$ that takes the task context~$\text{x}_t$, candidate image $\hat{\text{img}}_{t,i}$, and concept scores~$\hat{\text{s}}_{t,i}$ as input and outputs an acceptance probability:
\begin{equation}
  \hat{p}_{t,i}
  = f_\theta(\text{x}_t, \hat{\text{img}}_{t,i}, \hat{\text{s}}_{t,i})
  \in [0,1].
\end{equation}
The payment head is supervised using the binary payment labels
$y_{t,i} \in \{0,1\}$ obtained during annotation:
\begin{equation}
\resizebox{0.895\linewidth}{!}{$
\mathcal{L}_{\text{pay}}
  = \mathbb{E}_{t,i}\!\left[
      - y_{t,i}\log \hat{p}_{t,i}
      - (1-y_{t,i})\log(1-\hat{p}_{t,i})
    \right],
$}
\end{equation}
which encourages the predicted acceptance probability to align with empirical human payment decisions.

Concretely, we treat $\hat{p}_{t,i}$ as the acceptance probability of candidate image $i$ for task $t$, aggregate these probabilities within each task to obtain a task-level acceptance rate, and then combine this rate with the task price to compute each model’s total payment under our payment model.

\section{Experiments and Analysis}
\label{sec:Experiments_and_Results}

\begin{table*}[t]
    \centering
    \scriptsize
    \setlength{\tabcolsep}{2pt}
    \renewcommand{\arraystretch}{0.85}
    \begin{tabular}{lcccccccccc}
        \toprule
        \multirow{2}{*}{\textbf{Model}} &
        \multicolumn{4}{c}{\textbf{Total}} &
        \multicolumn{2}{c}{\textbf{Portrait}} &
        \multicolumn{2}{c}{\textbf{Product}} &
        \multicolumn{2}{c}{\textbf{Digital}} \\
        \cmidrule(lr){2-5}\cmidrule(lr){6-7}\cmidrule(lr){8-9}\cmidrule(lr){10-11}
        & \textbf{Rev (\$k)} & \textbf{Share (\%)} &
          \textbf{Task Acc. (\%)} & \textbf{Deliv. Acc. (\%)} &
          \textbf{Rev (\$k)} & \textbf{Share (\%)} &
          \textbf{Rev (\$k)} & \textbf{Share (\%)} &
          \textbf{Rev (\$k)} & \textbf{Share (\%)} \\
        \midrule
        \multicolumn{11}{c}{\emph{Closed-Source Models}} \\
        \midrule
        Gemini-Banana      & 109.27 & 37.0 & 29.91 & 34.56 & 2.08 & 13.6 & 50.72 & 34.5 & 56.48 & 42.5 \\
        Gemini-Banana-Pro  & \cellcolor{rankFirst}243.98 & \cellcolor{rankFirst}82.7 & \cellcolor{rankFirst}70.50 & \cellcolor{rankFirst}79.88 & 3.90 & 25.5 & \cellcolor{rankFirst}122.60 & \cellcolor{rankFirst}83.4 & \cellcolor{rankFirst}117.48 & \cellcolor{rankFirst}88.5 \\
        GPT-Image-1        & \cellcolor{rankSecond}148.32 & \cellcolor{rankSecond}50.3 & \cellcolor{rankSecond}48.22 & \cellcolor{rankSecond}46.38 & \cellcolor{rankThird}7.38 & \cellcolor{rankThird}48.3 & \cellcolor{rankThird}82.95 & \cellcolor{rankThird}56.4 & \cellcolor{rankThird}57.99 & \cellcolor{rankThird}43.7 \\
        GPT-Image-1-Mini   & \cellcolor{rankThird}113.04 & \cellcolor{rankThird}38.3 & 32.36 & 37.62 & 0.17 & 1.1 & \cellcolor{rankSecond}87.95 & \cellcolor{rankSecond}59.8 & 24.93 & 18.8 \\
        Ideogram-v3        & 80.53 & 27.3 & 25.89 & 31.67 & 1.01 & 6.6 & 20.94 & 14.2 & \cellcolor{rankSecond}58.57 & \cellcolor{rankSecond}44.1 \\
        Imagen-4.0         & 111.58 & 37.8 & 31.03 & 28.10 & 4.03 & 26.3 & 70.46 & 47.9 & 37.09 & 27.9 \\
        Kling-v2           & 96.15 & 32.6 & \cellcolor{rankThird}40.22 & 28.85 & \cellcolor{rankFirst}9.79 & \cellcolor{rankFirst}64.0 & 39.33 & 26.8 & 47.02 & 35.4 \\
        MJ-Relax-Edits     & 78.81 & 26.7 & 22.43 & 25.90 & 1.10 & 7.2 & 30.93 & 21.0 & 46.78 & 35.2 \\
        MJ-Relax-Imagine   & 88.08 & 29.9 & 37.66 & 28.01 & \cellcolor{rankSecond}8.82 & \cellcolor{rankSecond}57.7 & 48.20 & 32.8 & 31.06 & 23.4 \\
        SeedEdit-3.0-i2i   & 1.81 & 0.6 & 6.98 & 6.77 & 0.00 & 0.0 & 0.67 & 0.5 & 1.14 & 0.9 \\
        Seedream-3.0-t2i   & 105.21 & 35.7 & 31.40 & \cellcolor{rankThird}41.54 & 0.95 & 6.2 & 71.46 & 48.6 & 32.80 & 24.7 \\
        Seedream-4.0       & 65.95 & 22.4 & 31.12 & 40.18 & 2.44 & 15.9 & 46.45 & 31.6 & 17.06 & 12.8 \\
        \midrule
        \multicolumn{11}{c}{\emph{Open-Source Models}} \\
        \midrule
        FLUX-1.1-Pro       & 30.53 & 10.3 & 10.00 & 8.85 & 0.33 & 2.2 & 3.30 & 2.2 & 26.91 & 20.3 \\
        FLUX-Kontext-Pro   & 53.14 & 18.0 & 16.47 & 14.18 & 0.21 & 1.4 & 20.42 & 13.9 & 32.51 & 24.5 \\
        Qwen-Image         & 75.50 & 25.6 & 24.95 & 30.40 & 1.73 & 11.3 & 31.58 & 21.5 & 42.19 & 31.8 \\
        SD-3.5-Large       & 16.99 & 5.8 & 8.60 & 6.40 & 2.33 & 15.2 & 5.33 & 3.6 & 9.33 & 7.0 \\
        \bottomrule
    \end{tabular}
    \caption{Model performance on ServImage under the standard settlement scenario.
    Revenue, Rev (\$k), is the total contract value earned by a model, and
    Share is its fraction of the overall contract value (\$295k), while category shares are computed within each category.
    Task Acceptance and Deliverable Acceptance are the
    proportions of tasks and deliverables approved according to human payment decisions ($y_{t,i}$).
    The last six columns report revenue and share by Portrait, Product, and
    Digital categories. All monetary values are in thousand USD.
    \textbf{Color:}
    \legendbox{rankFirst}\,1st
    \ \legendbox{rankSecond}\,2nd 
    \ \legendbox{rankThird}\,3rd. 
    }
    \label{tab:main_results}
\end{table*}

\subsection{Setups}
We evaluate 16 image models, including 12 proprietary and 4 open-source models, for text-to-image generation and image editing. On the commercial side, we include OpenAI’s GPT-5-Image family, including gpt-5-image and gpt-5-image-mini~\cite{openai2025gptimage1,openai2025gpt5imagemini}; Google’s Gemini Nano Banana, Banana Pro, and Imagen-4.0~\cite{google2025gemini25flashimage,google2025imagen4api}; multiple Doubao systems, including Seedream 4.0 and Seedream 3.0-t2i for text-to-image generation~\cite{seedream2025seedream,gao2025seedream}, and SeedEdit 3.0-i2i for image-to-image editing~\cite{wang2025seededit}; Kuaishou’s Kling-v2~\cite{kuaishou2025kolors2image}; and models from professional design platforms such as Ideogram-v3~\cite{ideogram2025v3} and Midjourney’s Imagine and Edit pipelines~\cite{midjourney2024varyregion}. We also include the MJ-Relax-Edits and MJ-Relax-Imagine modes from Midjourney. On the open-source side, we include FLUX\text{-}1.1\text{-}Pro~\cite{blackforestlabs2024flux11pro}
and FLUX.1\text{-}Kontext\text{-}Pro~\cite{blackforestlabs2025fluxkontextpro},
Qwen\text{-}Image~\cite{wu2025qwenimage},
and Stable\text{-}Diffusion\text{-}3.5\text{-}Large~\cite{stabilityai2024sd35large}.
We use OpenAI's GPT\text{-}5\text{-}mini as an automatic judge to compute the BRF, VEQ, and CNS scores defined in Sec.~\ref{sec:ServImageScore}.

\subsection{Main Experiments}
\label{sec:Main_Experiments}
Under the standard settlement scenario, we evaluate 16 image models on ServImage and report their economic outcomes in Table~\ref{tab:main_results}.
Overall, the top proprietary models capture most of the economic value, for example, Gemini-Banana-Pro earns \$243.98k (82.7\% share) and GPT-Image-1 earns \$148.32k (50.3\%), while the strongest open-source baseline, Qwen-Image, obtains a markedly smaller payout of \$75.50k (25.6\%).
The best-performing model, Gemini-Banana-Pro, achieves the highest total revenue of \$243.98k and captures the largest contract share of 82.7\% out of the overall \$295k budget.
In contrast, the strongest open-source baseline, Qwen-Image, earns \$75.50k with a 25.6\% share, indicating a substantial gap in both maximum attainable revenue and market share under the same settlement rule.

Across the three fine-grained categories in ServImage, we observe clear leaders in economic outcomes. Kling-v2 leads Portrait with \$9.79k revenue (64.0\% within-category share), narrowly followed by MJ-Relax-Imagine (57.7\%). Gemini-Banana-Pro ranks first in Product with \$122.60k (83.4\%), with GPT-Image-1-Mini second (59.8\%), and also dominates Digital with \$117.48k (88.5\%), followed by Ideogram-v3 (44.1\%).
Despite these leaders, no single model achieves consistently high share across all categories, suggesting substantial room for improvement.
Table~\ref{tab:main_results} uses human accept/reject labels as ground truth; consequently, Table~\ref{tab:main_results_auto_by_serviamgemodel} reports an automatic proxy estimated from ServImageModel-predicted payment probabilities.
Additionally, the failure examples and corresponding analysis are provided in Appendix~\ref{app:Failure_Analyze}.

\begin{table}[h]
  \centering
  \small
  \setlength{\tabcolsep}{8.5pt}
  \renewcommand{\arraystretch}{0.8}
  \begin{tabular}{lcc}
    \toprule
    \textbf{Model} & \textbf{Rev (\$k)} & \textbf{Rank} \\
    \midrule
    Gemini 2.5 Flash               & \cellcolor{rankSecond}78.9 & \cellcolor{rankSecond}2 \\
    Gemini-Banana-Pro              & \cellcolor{rankFirst}113.7 & \cellcolor{rankFirst}1 \\
    GPT-5 Image                    & 10.1 & 5 \\
    GPT-5 Image Mini               &  5.8 & 6 \\
    imagen-4.0                     &  4.1 & 7 \\
    Seedream 4.0                   &  0.0 & 15 \\
    Seedream 3.0-t2i               &  0.3 & 13 \\
    SeedEdit 3.0-i2i               &  0.0 & 16 \\
    Kling-v2                       &  3.2 & 9 \\
    ideogram-v3                    &  0.9 & 10 \\
    MJ-Relax-Imagine               &  3.5 & 8 \\
    MJ-Relax-Edits                 &  0.4 & 11 \\
    \midrule
    \multicolumn{3}{c}{\textbf{Open-Source Models}} \\
    \midrule
    stable-diffusion-3.5-large     &  0.1 & 14 \\
    Flux Pro                       & 35.7 & 4 \\
    flux-kontext-pro               & \cellcolor{rankThird}37.9 & \cellcolor{rankThird}3 \\
    Qwen-Image                     &  0.3 & 12 \\
    \bottomrule
  \end{tabular}
    \caption{Crowdsourcing competition earnings with winner-takes-all acceptance. We report total value earned Rev which means Revenue, and Rank.
    \textbf{Color:}
    \legendbox{rankFirst}\,1st
    \ \legendbox{rankSecond}\,2nd
    \ \legendbox{rankThird}\,3rd.
  }
 \label{tab:crowd_competition}
\end{table}

\subsection{Crowdsourcing Competition}
The settlement scenario above assumes that a single provider is deployed at a
time: each model is evaluated on the full test split and earns revenue from all
tasks it solves.
In practice, commercial crowdsourcing platforms list a task to many providers, and only the best submission is paid.
To capture this more competitive regime, we construct a crowdsourcing competition setting in which all models are run on the same tasks, and for each task, we select the winner by the number of deliverables accepted by human payment labels, ties are broken by the summed ServImageScore (BRF/VEQ/CNS).
The full task revenue is then assigned to this ``winner'' model and all other
models receive zero payment for that task, resulting in a winner-takes-all
allocation of the total contract value, as shown in Table~\ref{tab:crowd_competition}.
In this section, a task is considered successful if and only if all its required deliverables are accepted (paid) according to the human payment-decision labels.

In the crowdsourcing competition setting, earnings are highly concentrated: a small number of top models capture a much larger portion of the total revenue, while many others earn little or even zero. However, the outcome is not completely one-sided. For example, Gemini-Banana-Pro ranks first with \$113.7k, but Gemini 2.5 Flash remains close behind at \$78.9k, and an open-source model (flux-kontext-pro) also achieves a strong result with \$37.9k.

\subsection{Cost Reduction Analysis}
The revenue analysis in Sec.~\ref{sec:Main_Experiments} quantifies each model’s earnings under our settlement rule.
In practice, platform operators and clients are also interested in a complementary question. If a model is deployed as the first stage of the pipeline and human designers only redo failed tasks, how much outsourcing cost can be saved, and how efficient is the model-first pipeline in terms of return per dollar spent?

We capture this economic perspective using model-level aggregate quantities.
For a given model $m$, we define
{\footnotesize\boldmath\color{black} $B = \sum_{t \in \mathcal{T}} \mathrm{Price}(t)$},
{\footnotesize\boldmath\color{black} $S_c(m) = 1 - \dfrac{\mathrm{Cost}_{\mathrm{API}}(m)}{B}
- \dfrac{1}{B} \sum_{t \in \mathcal{T}} \mathrm{Price}(t)\bigl(1 - \mathrm{Success}_c(m,t)\bigr)$},
and the \emph{Contribution Ratio}
{\footnotesize\boldmath\color{black}
$R_c(m)
= \dfrac{\sum_{t \in \mathcal{T}} \mathrm{Price}(t)\,\mathrm{Success}_c(m,t)}
{\mathrm{Cost}_{\mathrm{API}}(m) + \sum_{t \in \mathcal{T}} \mathrm{Price}(t)\bigl(1 - \mathrm{Success}_c(m,t)\bigr)}$}.
Here $S_c(m)$ is the overall cost savings relative to the human-only baseline, while $R_c(m)$ measures the return per dollar spent on the model API and freelancer rework under scenario $c$.
In the main text we report them as Cost Savings (\%) and Contribution Ratio. 
Per-call API price assumptions are listed in Appendix Table~\ref{tab:model_api_prices}.

Empirically, the best-performing model, MJ-Relax-Edits, achieves cost savings of $S_c(m) \approx 81.7\%$ and a contribution ratio of $R_c(m) \approx $, indicating that each dollar spent on API usage and human rework yields about \$4.4581 of end-to-end contract value handled by the model-first pipeline under realistic settlement rules.
Appendix Table~\ref{tab:cost_reduction} reports the full cost reduction results.

\subsection{Evaluation Method Comparison}

To evaluate ServImageScore in a payment-decision setting, we compare it with seven metrics in three categories.
Generation-specific metrics (OmniGenBench~\cite{wang2025omnigenbench}, OneIG-Bench~\cite{chang2025oneig}) focus on text-to-image and do not fit source-constrained editing.
Editing-specific metrics (VIEScore~\cite{ye2025imgedit}, RISEBench~\cite{zhao2025envisioning}) target editing but not source-free generation.
General-purpose metrics include ICE-Bench~\cite{pan2025ice}, I2EBench~\cite{ma2024i2ebench}, and LMM4Edit.

\begin{figure}[tb]
    \centering
    \includegraphics[width=\linewidth]{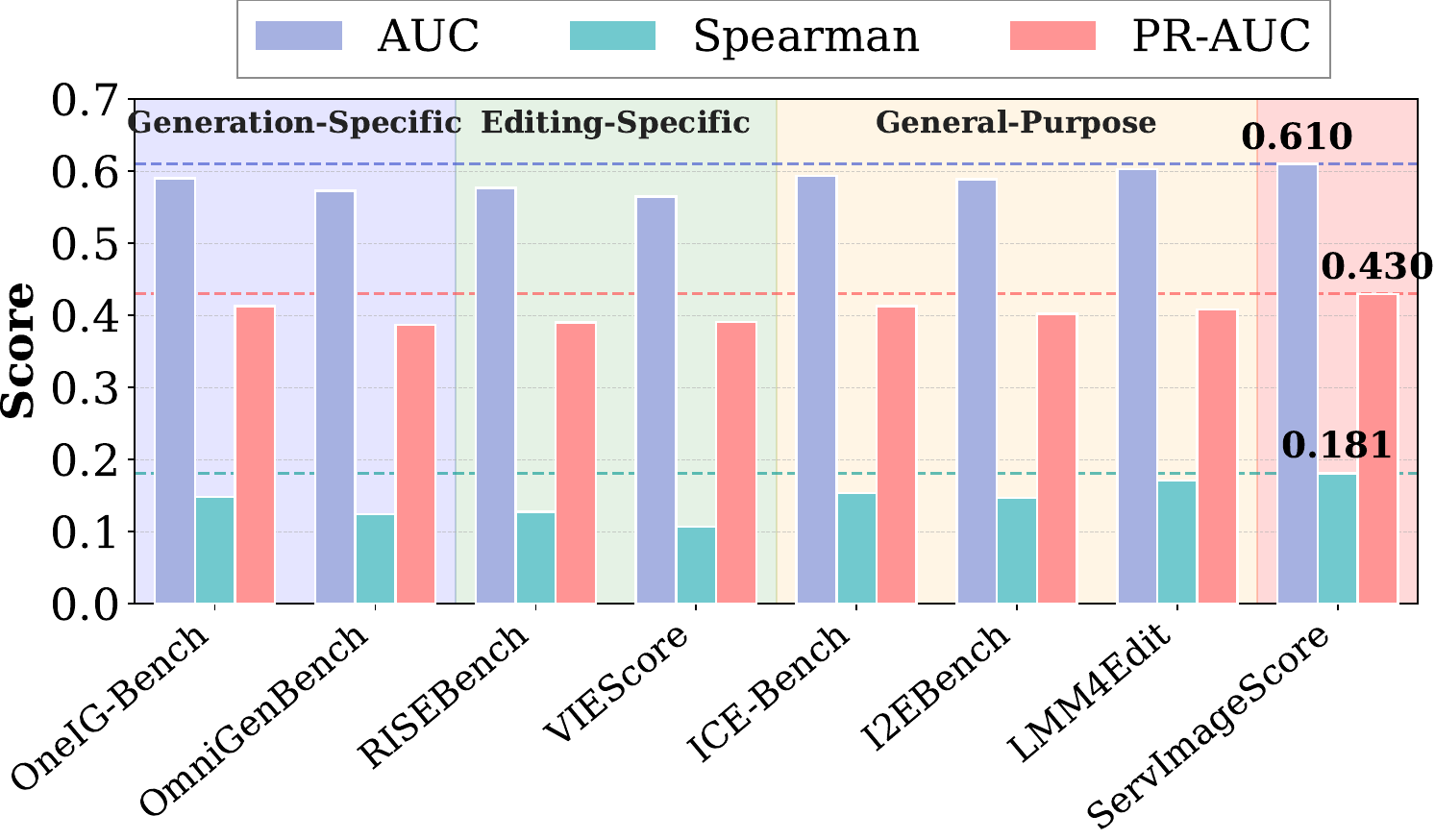}
    \caption{Metric comparison on the test set. Bars show AUC, Spearman, and PR-AUC for seven metrics and ServImageScore, higher is better.}
    \label{fig:metric_comparison}
\end{figure}

We examine whether automatic metrics can approximate human payment decisions.
We apply each metric to the same subset and evaluate its alignment with human payment labels using three criteria:
(i) AUC, the area under the ROC curve for binary payment outcomes;
(ii) Spearman, the rank correlation between metric scores and payment labels; and
(iii) PR-AUC, the area under the precision--recall curve for payment prediction.
Across all three metric groups, existing baselines show lower Spearman and PR-AUC than ServImageScore, suggesting weaker alignment with commercial acceptance.
As shown in Fig.~\ref{fig:metric_comparison}, ServImageScore performs best on all three criteria.

Additionally, the ServImageScore of different models, together with the judge robustness and reproducibility results relative to human evaluation, are provided in Appendix~\ref{app:Judge_Robustness_and_Reproducibility}, where the analysis shows that ServImageScore is robust.

\subsection{ServImageModel Analysis}

\paragraph{Comparison with General-purpose VLMs.}
Fig.~\ref{fig:workflow}(b) compares the payment-decision accuracy (\%) of general-purpose VLM baselines and ServImageModel on the test split.
Among all general-purpose baselines, our task-specific models substantially outperform all baselines: ServImageModel$_{\text{4B-Think}}$ reaches 82.00\%.

\begin{table}[h]
\centering
\footnotesize
\setlength{\tabcolsep}{3pt}
\renewcommand{\arraystretch}{0.8}
\begin{tabular}{lcccc}
\toprule
\textbf{Base Model} & \textbf{CBM} & \textbf{LoRA (v)} & \textbf{LoRA (l)} & \textbf{Acc.} \\
\midrule
    Qwen-4B-Think   &            &            &            & 72.12\% \\
    Qwen-4B-Think   & \checkmark &            &            & 78.30\% \\
    Qwen-4B-Think   & \checkmark & \checkmark &            & 79.21\% \\
    \rowcolor{gray!15}
    Qwen-4B-Think   & \checkmark & \checkmark & \checkmark & \textbf{82.00\%} \\
    \midrule
    Qwen-4B-Instr.  & \checkmark & \checkmark & \checkmark & 79.67\% \\
\bottomrule
\end{tabular}
\caption{Ablation study on the CBM interface, LoRA strategies, and Qwen3-VL backbones. We report payment-decision accuracy on the test split.}
\label{tab:ablation_cbm_lora_backbone}
\end{table}

\paragraph{Ablation Study.}
To further understand why ServImageModel. achieves large gains in Fig.~\ref{fig:workflow}(b), we ablate three components of the payment model: the CBM interface, LoRA adapters on the vision and language towers, and the choice of Qwen3-VL backbone, as shown in Table~\ref{tab:ablation_cbm_lora_backbone}. 
Using Qwen-4B-Think as the base model, the plain baseline achieves 72.12\% accuracy. Adding LoRA on both vision and language yields the best performance of 82.00\%.
Together, these two findings demonstrate the effectiveness of the ServImageModel framework for predicting payment decisions.

\section{Conclusion}
\label{sec:Conclusion}
We present ServImage, a benchmark for evaluating image generation and editing models on real-world commercial design tasks and payment decisions. By connecting model outputs with task prices and human acceptance outcomes, and introducing ServImageScore to decompose commercial acceptability into baseline requirement fulfillment, visual execution quality, and commercial necessity satisfaction, ServImage moves beyond conventional metrics and better reflects why people pay. We hope ServImage will encourage market-aware evaluation and foster vision models that are more economically aligned.
\section*{Limitations}
ServImageBench is constructed from paid design tasks collected from two Chinese crowdsourcing platforms. As a result, task distributions, aesthetic preferences, and pricing norms may vary across regions, languages, and procurement settings. As a first look at linking image generation and editing performance to real commercial payment outcomes, we view this benchmark as an exploratory starting point, and future efforts can broaden coverage to more regions, languages, and enterprise-level design scenarios to improve generalizability.

In addition, Commercial acceptability is currently operationalized as a binary payment decision, which provides a clear and reproducible economic signal consistent with the settlement rules of real crowdsourcing platforms. However, real-world design workflows are typically more iterative and fine-grained, involving partial acceptance, repeated revisions, and negotiated outcomes. Under such conditions, commercial value depends not only on initial output quality but also on a model’s stability, editability, and robustness across multiple rounds of interaction. Consequently, the current benchmark does not fully capture the iterative nature of commercial design or its broader economic implications. Future work may extend this formulation by incorporating revision-based workflows, multi-turn evaluation, and more continuous representations of commercial outcomes.

Finally, Prior public evaluations largely use human preference signals or subjective ratings, with emerging efforts incorporating online utility metrics such as engagement or CTR. In contrast, we did not find publicly available benchmarks that map VLM-based quality scores for image generation/editing to real payment/settlement outcomes from commercial design orders. Accordingly, we treat our benchmark as an exploratory starting point, and future work can enhance robustness and temporal stability via multi-judge ensembles, cross-model calibration, and selective human verification.

\section*{Ethics Statement}
ServImage is derived from real commercial design orders that may include user-provided reference materials or source images, so privacy and responsible data handling are essential. 
We anonymized all released data and removed personally identifying information, and we restricted or redacted potentially sensitive assets whenever sharing them was not appropriate.
Human payment labels are produced through a structured annotation process, and annotators participate voluntarily under clear guidelines.
Finally, the benchmark’s automated scoring uses external model services only for evaluation and should be conducted in compliance with provider policies, with no attempt to bypass safety measures or misuse protected content.

\section*{Acknowledgements}
We thank the anonymous reviewers and the area chair for their constructive comments. We also thank our mentors and colleagues from MBZUAI for their support and help.

{
    \small
    \bibliography{main}

@String(CVPR= {IEEE Conf. Comput. Vis. Pattern Recog.})

@String(CVPR  = {CVPR})

@article{miserendino2025swe,
  title={SWE-Lancer: Can Frontier LLMs Earn \$1 Million from Real-World Freelance Software Engineering?},
  author={Miserendino, Samuel and Wang, Michele and Patwardhan, Tejal and Heidecke, Johannes},
  journal={arXiv preprint arXiv:2502.12115},
  year={2025}
}

@article{wu2025kris,
  title={KRIS-Bench: Benchmarking Next-Level Intelligent Image Editing Models},
  author={Wu, Yongliang and Li, Zonghui and Hu, Xinting and Ye, Xinyu and Zeng, Xianfang and Yu, Gang and Zhu, Wenbo and Schiele, Bernt and Yang, Ming-Hsuan and Yang, Xu},
  journal={arXiv preprint arXiv:2505.16707},
  year={2025}
}

@article{zhao2025envisioning,
  title={Envisioning beyond the pixels: Benchmarking reasoning-informed visual editing},
  author={Zhao, Xiangyu and Zhang, Peiyuan and Tang, Kexian and Zhu, Xiaorong and Li, Hao and Chai, Wenhao and Zhang, Zicheng and Xia, Renqiu and Zhai, Guangtao and Yan, Junchi and others},
  journal={arXiv preprint arXiv:2504.02826},
  year={2025}
}

@inproceedings{wanglmm4lmm,
  title={LMM4LMM: Benchmarking and Evaluating Large-multimodal Image Generation with LMMs},
  author={Wang, Jiarui and Duan, Huiyu and Zhao, Yu and Wang, Juntong and Zhai, Guangtao and Min, Xiongkuo},
  booktitle={Proceedings of the IEEE/CVF International Conference on Computer Vision},
  year={2025}
}

@article{wang2025omnigenbench,
  title={OmniGenBench: A Benchmark for Omnipotent Multimodal Generation across 50+ Tasks},
  author={Wang, Jiayu and Jiao, Yang and Yu, Yue and Qian, Tianwen and Chen, Shaoxiang and Chen, Jingjing and Jiang, Yu-Gang},
  journal={arXiv preprint arXiv:2505.18775},
  year={2025}
}

@inproceedings{farshad2023scenegenie,
  title={Scenegenie: Scene graph guided diffusion models for image synthesis},
  author={Farshad, Azade and Yeganeh, Yousef and Chi, Yu and Shen, Chengzhi and Ommer, B{\"o}jrn and Navab, Nassir},
  booktitle={Proceedings of the IEEE/CVF International Conference on Computer Vision},
  pages={88--98},
  year={2023}
}

@inproceedings{zeng2024dilightnet,
  title={DiLightNet: Fine-grained lighting control for diffusion-based image generation},
  author={Zeng, Chong and Dong, Yue and Peers, Pieter and Kong, Youkang and Wu, Hongzhi and Tong, Xin},
  booktitle={ACM SIGGRAPH 2024 Conference Papers},
  pages={1--12},
  year={2024}
}

@article{ma2024i2ebench,
  title={I2ebench: A comprehensive benchmark for instruction-based image editing},
  author={Ma, Yiwei and Ji, Jiayi and Ye, Ke and Lin, Weihuang and Wang, Zhibin and Zheng, Yonghan and Zhou, Qiang and Sun, Xiaoshuai and Ji, Rongrong},
  journal={Advances in Neural Information Processing Systems},
  volume={37},
  pages={41494--41516},
  year={2024}
}

@article{ye2025imgedit,
  title={Imgedit: A unified image editing dataset and benchmark},
  author={Ye, Yang and He, Xianyi and Li, Zongjian and Lin, Bin and Yuan, Shenghai and Yan, Zhiyuan and Hou, Bohan and Yuan, Li},
  journal={arXiv preprint arXiv:2505.20275},
  year={2025}
}

@article{xu2025lmm4edit,
  title={Lmm4edit: Benchmarking and evaluating multimodal image editing with lmms},
  author={Xu, Zitong and Duan, Huiyu and Liu, Bingnan and Ma, Guangji and Wang, Jiarui and Yang, Liu and Gao, Shiqi and Wang, Xiaoyu and Wang, Jia and Min, Xiongkuo and others},
  journal={arXiv preprint arXiv:2507.16193},
  year={2025}
}

@article{pu2025picabench,
  title={PICABench: How Far Are We from Physically Realistic Image Editing?},
  author={Pu, Yuandong and Zhuo, Le and Han, Songhao and Xing, Jinbo and Zhu, Kaiwen and Cao, Shuo and Fu, Bin and Liu, Si and Li, Hongsheng and Qiao, Yu and others},
  journal={arXiv preprint arXiv:2510.17681},
  year={2025}
}

@article{seedream2025seedream,
  title={Seedream 4.0: Toward next-generation multimodal image generation},
  author={Seedream, Team and Chen, Yunpeng and Gao, Yu and Gong, Lixue and Guo, Meng and Guo, Qiushan and Guo, Zhiyao and Hou, Xiaoxia and Huang, Weilin and Huang, Yixuan and others},
  journal={arXiv preprint arXiv:2509.20427},
  year={2025}
}

@misc{openai2025gptimage1,
  author       = {OpenAI},
  title        = {GPT-Image-1: OpenAI's Multimodal Image Generation Model},
  year         = {2025},
  howpublished = {\url{https://platform.openai.com/docs/models/gpt-image-1}},
  note         = {Accessed: 2025-05-08}
}

@misc{openai2025gpt5imagemini,
  author       = {OpenAI},
  title        = {GPT-5 Image Mini: Cost-Efficient Multimodal Image Generation Model},
  year         = {2025},
  month        = {October},
  howpublished = {\url{https://platform.openai.com/docs/models/gpt-image-1-mini}},
  note         = {Accessed: 2025-11-03}
}

@misc{google2025imagen4api,
  author       = {Google Cloud},
  title        = {Imagen 4 Generate API Reference},
  year         = {2025},
  month        = {October},
  howpublished = {\url{https://cloud.google.com/vertex-ai/generative-ai/docs/models/imagen/4-0-generate-001}},
  note         = {Accessed: 2025-11-06}
}

@misc{google2025gemini25flashimage,
  author       = {Google DeepMind},
  title        = {Gemini 2.5 Flash Image (Nano Banana)},
  year         = {2025},
  month        = {August},
  howpublished = {\url{https://developers.googleblog.com/en/introducing-gemini-2-5-flash-image/}}
}

@article{wang2025seededit,
  title={SeedEdit 3.0: Fast and High-Quality Generative Image Editing},
  author={Wang, Peng and Shi, Yichun and Lian, Xiaochen and Zhai, Zhonghua and Xia, Xin and Xiao, Xuefeng and Huang, Weilin and Yang, Jianchao},
  journal={arXiv preprint arXiv:2506.05083},
  year={2025}
}

@article{gao2025seedream,
  title={Seedream 3.0 technical report},
  author={Gao, Yu and Gong, Lixue and Guo, Qiushan and Hou, Xiaoxia and Lai, Zhichao and Li, Fanshi and Li, Liang and Lian, Xiaochen and Liao, Chao and Liu, Liyang and others},
  journal={arXiv preprint arXiv:2504.11346},
  year={2025}
}

@inproceedings{li2024genai,
  title={Genai-bench: A holistic benchmark for compositional text-to-visual generation},
  author={Li, Baiqi and Lin, Zhiqiu and Pathak, Deepak and Li, Jiayao Emily and Xia, Xide and Neubig, Graham and Zhang, Pengchuan and Ramanan, Deva},
  booktitle={Synthetic Data for Computer Vision Workshop@ CVPR 2024},
  year={2024}
}

@article{huang2025t2i,
  title={T2i-compbench++: An enhanced and comprehensive benchmark for compositional text-to-image generation},
  author={Huang, Kaiyi and Duan, Chengqi and Sun, Kaiyue and Xie, Enze and Li, Zhenguo and Liu, Xihui},
  journal={IEEE Transactions on Pattern Analysis and Machine Intelligence},
  year={2025},
  publisher={IEEE}
}

@article{chang2025oneig,
  title={OneIG-Bench: Omni-dimensional Nuanced Evaluation for Image Generation},
  author={Chang, Jingjing and Fang, Yixiao and Xing, Peng and Wu, Shuhan and Cheng, Wei and Wang, Rui and Zeng, Xianfang and Yu, Gang and Chen, Hai-Bao},
  journal={arXiv preprint arXiv:2506.07977},
  year={2025}
}

@article{basu2023editval,
  title={Editval: Benchmarking diffusion based text-guided image editing methods},
  author={Basu, Samyadeep and Saberi, Mehrdad and Bhardwaj, Shweta and Chegini, Atoosa Malemir and Massiceti, Daniela and Sanjabi, Maziar and Hu, Shell Xu and Feizi, Soheil},
  journal={arXiv preprint arXiv:2310.02426},
  year={2023}
}

@article{sun2025ie,
  title={Ie-bench: Advancing the measurement of text-driven image editing for human perception alignment},
  author={Sun, Shangkun and Qu, Bowen and Liang, Xiaoyu and Fan, Songlin and Gao, Wei},
  journal={arXiv preprint arXiv:2501.09927},
  year={2025}
}

@article{pan2025ice,
  title={Ice-bench: A unified and comprehensive benchmark for image creating and editing},
  author={Pan, Yulin and He, Xiangteng and Mao, Chaojie and Han, Zhen and Jiang, Zeyinzi and Zhang, Jingfeng and Liu, Yu},
  journal={arXiv preprint arXiv:2503.14482},
  year={2025}
}

@inproceedings{hu2024instruct,
  title={Instruct-imagen: Image generation with multi-modal instruction},
  author={Hu, Hexiang and Chan, Kelvin CK and Su, Yu-Chuan and Chen, Wenhu and Li, Yandong and Sohn, Kihyuk and Zhao, Yang and Ben, Xue and Gong, Boqing and Cohen, William and others},
  booktitle={Proceedings of the IEEE/CVF conference on computer vision and pattern recognition},
  pages={4754--4763},
  year={2024}
}

@misc{kuaishou2025kolors2image,
  author       = {Kuaishou Technology},
  title        = {KOLORS 2.0: Standalone Image Module for Advanced Image Synthesis},
  year         = {2025},
  month        = {April},
  howpublished = {\url{https://app.klingai.com/global/}},
  note         = {Image generation component of Kling 2.0}
}

@misc{ideogram2025v3,
  author       = {Ideogram},
  title        = {Ideogram V3: Advanced AI Image Generation and Editing Model},
  year         = {2025},
  month        = {March},
  day          = {26},
  howpublished = {\url{https://docs.ideogram.ai/}},
  note         = {Released March 26, 2025}
}

@misc{midjourney2024varyregion,
  author       = {Midjourney},
  title        = {Vary Region: Advanced Image Inpainting Tool},
  year         = {2024},
  month        = {August},
  howpublished = {\url{https://docs.midjourney.com/hc/en-us/articles/32794723105549-Vary-Region}}
}

@phdthesis{1, title={Learning visual representations without human supervision}, url={http://dx.doi.org/10.32657/10356/171772}, DOI={10.32657/10356/171772}, school={Nanyang Technological University}, author={Xie, Jiahao} }

@inproceedings{rombach2022high,
  title={High-resolution image synthesis with latent diffusion models},
  author={Rombach, Robin and Blattmann, Andreas and Lorenz, Dominik and Esser, Patrick and Ommer, Bj{\"o}rn},
  booktitle={Proceedings of the IEEE/CVF conference on computer vision and pattern recognition},
  pages={10684--10695},
  year={2022}
}

@article{nichol2021glide,
  title={Glide: Towards photorealistic image generation and editing with text-guided diffusion models},
  author={Nichol, Alex and Dhariwal, Prafulla and Ramesh, Aditya and Shyam, Pranav and Mishkin, Pamela and McGrew, Bob and Sutskever, Ilya and Chen, Mark},
  journal={arXiv preprint arXiv:2112.10741},
  year={2021}
}

@inproceedings{brooks2023instructpix2pix,
  title={Instructpix2pix: Learning to follow image editing instructions},
  author={Brooks, Tim and Holynski, Aleksander and Efros, Alexei A},
  booktitle={Proceedings of the IEEE/CVF conference on computer vision and pattern recognition},
  pages={18392--18402},
  year={2023}
}

@article{zhang2023magicbrush,
  title={Magicbrush: A manually annotated dataset for instruction-guided image editing},
  author={Zhang, Kai and Mo, Lingbo and Chen, Wenhu and Sun, Huan and Su, Yu},
  journal={Advances in Neural Information Processing Systems},
  volume={36},
  pages={31428--31449},
  year={2023}
}

@inproceedings{radford2021learning,
  title={Learning transferable visual models from natural language supervision},
  author={Radford, Alec and Kim, Jong Wook and Hallacy, Chris and Ramesh, Aditya and Goh, Gabriel and Agarwal, Sandhini and Sastry, Girish and Askell, Amanda and Mishkin, Pamela and Clark, Jack and others},
  booktitle={International conference on machine learning},
  pages={8748--8763},
  year={2021},
  organization={PmLR}
}

@inproceedings{zhang2018unreasonable,
  title={The unreasonable effectiveness of deep features as a perceptual metric},
  author={Zhang, Richard and Isola, Phillip and Efros, Alexei A and Shechtman, Eli and Wang, Oliver},
  booktitle={Proceedings of the IEEE conference on computer vision and pattern recognition},
  pages={586--595},
  year={2018}
}

@article{wang2004image,
  title={Image quality assessment: from error visibility to structural similarity},
  author={Wang, Zhou and Bovik, Alan C and Sheikh, Hamid R and Simoncelli, Eero P},
  journal={IEEE transactions on image processing},
  volume={13},
  number={4},
  pages={600--612},
  year={2004},
  publisher={IEEE}
}

@article{chen2024mega,
  title={Mega-bench: Scaling multimodal evaluation to over 500 real-world tasks},
  author={Chen, Jiacheng and Liang, Tianhao and Siu, Sherman and Wang, Zhengqing and Wang, Kai and Wang, Yubo and Ni, Yuansheng and Zhu, Wang and Jiang, Ziyan and Lyu, Bohan and others},
  journal={arXiv preprint arXiv:2410.10563},
  year={2024}
}

@inproceedings{ryu2025towards,
  title={Towards Scalable Human-aligned Benchmark for Text-guided Image Editing},
  author={Ryu, Suho and Kim, Kihyun and Baek, Eugene and Shin, Dongsoo and Lee, Joonseok},
  booktitle={Proceedings of the Computer Vision and Pattern Recognition Conference},
  pages={18292--18301},
  year={2025}
}

@inproceedings{gu2025blendergym,
  title={BlenderGym: Benchmarking Foundational Model Systems for Graphics Editing},
  author={Gu, Yunqi and Huang, Ian and Je, Jihyeon and Yang, Guandao and Guibas, Leonidas},
  booktitle={Proceedings of the Computer Vision and Pattern Recognition Conference},
  pages={18574--18583},
  year={2025}
}

@article{peng2024dreambench++,
  title={Dreambench++: A human-aligned benchmark for personalized image generation},
  author={Peng, Yuang and Cui, Yuxin and Tang, Haomiao and Qi, Zekun and Dong, Runpei and Bai, Jing and Han, Chunrui and Ge, Zheng and Zhang, Xiangyu and Xia, Shu-Tao},
  journal={arXiv preprint arXiv:2406.16855},
  year={2024}
}

@inproceedings{zhang2025upme,
  title={UPME: An Unsupervised Peer Review Framework for Multimodal Large Language Model Evaluation},
  author={Zhang, Qihui and Ning, Munan and Liu, Zheyuan and Huang, Yue and Yang, Shuo and Wang, Yanbo and Ye, Jiayi and Chen, Xiao and Song, Yibing and Yuan, Li},
  booktitle={Proceedings of the Computer Vision and Pattern Recognition Conference},
  pages={9165--9174},
  year={2025}
}

@article{zhou2025draw,
  title={Draw ALL Your Imagine: A Holistic Benchmark and Agent Framework for Complex Instruction-based Image Generation},
  author={Zhou, Yucheng and Yuan, Jiahao and Wang, Qianning},
  journal={arXiv preprint arXiv:2505.24787},
  year={2025}
}

@inproceedings{wang2025complexbench,
  title={Complexbench-edit: Benchmarking complex instruction-driven image editing via compositional dependencies},
  author={Wang, Chenglin and Zhou, Yucheng and Wang, Qianning and Wang, Zhe and Zhang, Kai},
  booktitle={Proceedings of the 33rd ACM International Conference on Multimedia},
  pages={13391--13397},
  year={2025}
}

@article{patwardhan2025gdpval,
  title={Gdpval: Evaluating ai model performance on real-world economically valuable tasks},
  author={Patwardhan, Tejal and Dias, Rachel and Proehl, Elizabeth and Kim, Grace and Wang, Michele and Watkins, Olivia and Fishman, Sim{\'o}n Posada and Aljubeh, Marwan and Thacker, Phoebe and Fauconnet, Laurance and others},
  journal={arXiv preprint arXiv:2510.04374},
  year={2025}
}

@article{mazeika2025remote,
  title={Remote labor index: Measuring ai automation of remote work},
  author={Mazeika, Mantas and Gatti, Alice and Menghini, Cristina and Sehwag, Udari Madhushani and Singhal, Shivam and Orlovskiy, Yury and Basart, Steven and Sharma, Manasi and Peskoff, Denis and Lau, Elaine and others},
  journal={arXiv preprint arXiv:2510.26787},
  year={2025}
}

@article{xu2025manipshield,
  title={ManipShield: A Unified Framework for Image Manipulation Detection, Localization and Explanation},
  author={Xu, Zitong and Duan, Huiyu and Wang, Xiaoyu and Cai, Zhaolin and Zhang, Kaiwei and Hu, Qiang and Liu, Jing and Min, Xiongkuo and Zhai, Guangtao},
  journal={arXiv preprint arXiv:2511.14259},
  year={2025}
}

@misc{blackforestlabs2024flux11pro,
  author       = {Black Forest Labs},
  title        = {Announcing FLUX1.1 [pro] and the BFL API},
  year         = {2024},
  month        = {October},
  howpublished = {\url{https://bfl.ai/blog/24-10-02-flux}},
  note         = {Accessed: 2025-12-11}
}

@misc{blackforestlabs2025fluxkontextpro,
  author       = {Black Forest Labs},
  title        = {Introducing FLUX.1 Kontext and the BFL Playground},
  year         = {2025},
  month        = {May},
  howpublished = {\url{https://bfl.ai/blog/flux-1-kontext}},
  note         = {Accessed: 2025-12-11}
}

@misc{wu2025qwenimage,
  author       = {Chenfei Wu et al.},
  title        = {Qwen-Image Technical Report},
  year         = {2025},
  month        = {August},
  howpublished = {\url{https://arxiv.org/abs/2508.02324}},
  note         = {Accessed: 2025-12-11}
}

@misc{stabilityai2024sd35large,
  author       = {Stability AI},
  title        = {Introducing Stable Diffusion 3.5},
  year         = {2024},
  month        = {October},
  howpublished = {\url{https://stability.ai/news/introducing-stable-diffusion-3-5}},
  note         = {Accessed: 2025-12-11}
}

@inproceedings{jin2019pubmedqa,
  title={Pubmedqa: A dataset for biomedical research question answering},
  author={Jin, Qiao and Dhingra, Bhuwan and Liu, Zhengping and Cohen, William and Lu, Xinghua},
  booktitle={Proceedings of the 2019 conference on empirical methods in natural language processing and the 9th international joint conference on natural language processing (EMNLP-IJCNLP)},
  pages={2567--2577},
  year={2019}
}

@inproceedings{ku2024viescore,
  title={Viescore: Towards explainable metrics for conditional image synthesis evaluation},
  author={Ku, Max and Jiang, Dongfu and Wei, Cong and Yue, Xiang and Chen, Wenhu},
  booktitle={Proceedings of the 62nd Annual Meeting of the Association for Computational Linguistics (Volume 1: Long Papers)},
  pages={12268--12290},
  year={2024}
}

@article{yang2025agent,
  title={Agent-Based Anti-Jamming Techniques for UAV Communications in Adversarial Environments: A Comprehensive Survey},
  author={Yang, Jingpu and Cui, Mingxuan and Zhang, Hang and Ji, Fengxian and Lai, Zhengzhao and Wang, Yufeng},
  journal={arXiv preprint arXiv:2508.11687},
  year={2025}
}

@article{yang2026frequency,
  title={Frequency point game environment for uavs via expert knowledge and large language model},
  author={Yang, Jingpu and Zhang, Hang and Ji, Fengxian and Wang, Yufeng and Wang, Mingjie and Luo, Yizhe and Ding, Wenrui},
  journal={Drones},
  volume={10},
  number={2},
  pages={147},
  year={2026},
  publisher={MDPI}
}

@article{ji2025finestate,
  title={FineState-Bench: A Comprehensive Benchmark for Fine-Grained State Control in GUI Agents},
  author={Ji, Fengxian and Yang, Jingpu and Song, Zirui and Wang, Yuanxi and Cui, Zhexuan and Li, Yuke and Jiang, Qian and Fang, Miao and Chen, Xiuying},
  journal={arXiv preprint arXiv:2508.09241},
  year={2025}
}

@article{li2026m3mad,
  title={M3MAD-Bench: Are Multi-Agent Debates Really Effective Across Domains and Modalities?},
  author={Li, Ao and Zhang, Jinghui and Li, Luyu and Duan, Yuxiang and Gao, Lang and Chen, Mingcai and Qin, Weijun and Li, Shaopeng and Ji, Fengxian and Liu, Ning and others},
  journal={arXiv preprint arXiv:2601.02854},
  year={2026}
}

@inproceedings{zhang2025individuals,
  title={From Individuals to Crowds: Dual-Level Public Response Prediction in Social Media},
  author={Zhang, Jinghui and Wan, Kaiyang and Xu, Longwei and Li, Ao and Liu, Zongfang and Chen, Xiuying},
  booktitle={Proceedings of the 33rd ACM International Conference on Multimedia},
  pages={5903--5912},
  year={2025}
}

@article{liang2025vision,
  title={Vision Language Models Are Not (Yet) Spelling Correctors},
  author={Liang, Junhong and Zhang, Bojun},
  journal={arXiv preprint arXiv:2509.17418},
  year={2025}
}
}

\appendix

\FloatBarrier     
\onecolumn          
\section{Comparison with Existing Image Generation and Editing Benchmarks}
\label{sec:image_bench_comparison}

\vspace{-0.5em} 
\begin{center}
\begin{sideways}
\begin{minipage}{0.93\textheight} 
\centering

\scriptsize
\setlength{\tabcolsep}{3pt}
\renewcommand{\arraystretch}{1.25}

\begin{tabular}{%
    p{1.8cm}
    p{2.2cm}
    p{2.4cm}
    p{2.2cm}
    p{3.0cm}
    p{1.7cm}
    p{3.0cm}
    p{2.2cm}
}
\toprule
\textbf{Benchmark} &
\textbf{Domain} &
\textbf{Data source} &
\textbf{Data form} &
\textbf{Capabilities tested} &
\textbf{Monetary metric} &
\textbf{Evaluation method} &
\textbf{Judge source} \\
\midrule
ServImage &
Image generation and editing &
Paid real-world crowdsourcing tasks &
Commercial briefs and delivered images &
Business rules following, visual quality, set consistency, settlement decisions &
Yes (earnings and cost savings) &
BRF/VEQ/CNS scoring and cost-sensitive settlement rule &
GPT-5 with human validation \\
\midrule
GenAI-Bench~\cite{li2024genai} &
Text-to-image and video &
Real prompts from professional designers &
Prompts with human ratings and model outputs &
Compositional text-to-visual generation &
No &
Human mean-opinion scores and VQA-style metrics (VQAScore) &
Crowdsourced raters and MLLM-as-a-judge \\
OmniGenBench~\cite{wang2025omnigenbench} &
Text-to-image &
Curated reversible tasks and prompts &
Generated images across multiple categories &
Perception-centric and cognition-centric generation ability &
No &
Multi-dimensional automatic metrics and GPT-4o scoring &
MLLM-as-a-judge \\
T2I-CompBench++~\cite{huang2025t2i} &
Text-to-image &
Open-world compositional prompts &
Generated images with compositional structures &
Attribute and relation compositionality &
No &
CLIP and detection-based metrics plus MLLM-based evaluation &
Automatic metrics and MLLM-as-a-judge \\
OneIG-Bench~\cite{chang2025oneig} &
Text-to-image &
Curated prompts and annotations &
Prompts and generated images &
Subject alignment, text rendering, reasoning, stylization, diversity &
No &
Fine-grained multi-dimensional automatic scores &
Benchmark authors \\
\midrule
EDITVAL~\cite{basu2023editval} &
Text-guided image editing &
Images from MS-COCO with edit attributes &
Source--edit pairs with edit type labels &
Edit fidelity and content preservation across diverse edit types &
No &
Standardized VLM-based evaluation pipeline with human validation &
Pre-trained vision--language models and human study \\
VIEScore~\cite{ku2024viescore} &
Conditional image synthesis &
Aggregated datasets from seven conditional tasks &
Conditioned images and prompts or instructions &
General conditional image generation quality and explainable scoring &
No &
Single unified metric derived from MLLM responses &
GPT-4o and other multimodal language models \\
I2EBench~\cite{ma2024i2ebench} &
Instruction-based image editing &
More than 2{,}000 source images and 4{,}000 instructions &
Source--instruction--edit triples &
Sixteen dimensions of editing quality and alignment &
No &
Automated multi-dimensional evaluation aligned with user study &
Hybrid automatic metrics and user study \\
IE-Bench~\cite{sun2025ie} &
Text-driven image editing &
Collected source images and edit results &
Source--prompt--edit triples with MOS labels &
Perceptual quality and text--image consistency for editing &
No &
MOS-based image quality assessment and IE-QA metric &
Human mean-opinion scores and learned IQA model \\
RISEBench~\cite{zhao2025envisioning} &
Reasoning-informed visual editing &
Curated reasoning-aware editing cases &
Multi-step visual editing tasks with textual descriptions &
Temporal, causal, spatial, and logical reasoning in editing &
No &
LMM-as-a-judge pipeline with human checks &
GPT-4o and human raters \\
\midrule
ICE-Bench~\cite{pan2025ice} &
Image creating and editing &
Mixture of real and synthetic scenes &
Tasks in four categories and thirty-one fine-grained types &
Aesthetic quality, imaging quality, prompt following, consistency, controllability &
No &
Eleven automatic metrics including VLLM-QA for editing success &
Automatic metrics and MLLM-as-a-judge \\
Instruct-Imagen~\cite{hu2024instruct} &
Instructional image generation &
Multi-modal instruction data and curated datasets &
Image--text instruction pairs and evaluation sets &
Multi-modal instruction following and generalization &
No &
Human preference evaluation on multiple image generation datasets &
Human raters \\
LMM4Edit~\cite{xu2025lmm4edit} &
Text-guided image editing &
EBench-18K with edited images and human preferences &
Source--prompt--edit triples with MOS and question--answer pairs &
Perceptual quality, editing alignment, attribute preservation, task-specific QA &
No &
LMM4Edit metric learned from human preferences and QA signals &
Multimodal language model trained on MOS annotations \\
\bottomrule
\end{tabular}
\captionof{table}{Comparison of ServImage with existing image generation and editing benchmarks.
ServImage is the only benchmark that directly links technical scores to monetary outcomes
on paid commercial tasks, while prior work focuses on technical quality or human
preference without explicit settlement or earnings modeling.}
\label{tab:image_bench_comparison}
\end{minipage}
\end{sideways}
\end{center}

\clearpage
\twocolumn

\section{Cost Metrics: Definitions and Implementation Details}
\FloatBarrier
\subsection{API Price Assumptions}
\label{subsection:API_Price_Assumptions}
This subsection documents the unit API price assumptions for all evaluated
models, which serve as inputs to $\mathrm{Cost}_{\mathrm{API}}(m)$ in
Appendix~\ref{app:cost_metrics}. Unit: USD / call (i.e., the API cost
charged per model invocation/request in our implementation; equivalently, it can
be viewed as USD per generated/edited image when each call returns a single
final output).

\begin{table}[!htbp]
  \centering
  \footnotesize
  \setlength{\tabcolsep}{6pt}
  \renewcommand{\arraystretch}{1.15}
  \begin{tabular}{l r}
    \toprule
    \textbf{Model (API endpoint)} & \textbf{API price (USD / call)} \\
    \midrule
    gpt-image-1-mini              & 0.027542370 \\
    gpt-image-1                   & 0.137406021 \\
    gemini-2.5-flash-image-pro    & 0.028000000 \\
    gemini-2.5-flash-image        & 0.015947911 \\
    seedream-4.0                  & 0.027800000 \\
    seedream-3.0-t2i              & 0.036000000 \\
    seededit-3.0-i2i              & 0.041700000 \\
    kling-v2-images               & 0.027800000 \\
    ideogram-v3-text-to-image     & 0.030000000 \\
    mj-relax-imagine              & 0.019800000 \\
    mj-relax-edits                & 0.019728000 \\
    flux-1.1-pro                  & 0.040000000 \\
    flux-kontext-pro              & 0.040000000 \\
    qwen-text-to-image            & 0.034000000 \\
    stable-diffusion-3.5-large    & 0.065000000 \\
    imagen4                       & 0.040000000 \\
    \bottomrule
  \end{tabular}
    \caption{Per-call API prices (USD) assumed for each model in our experiments.}
  \label{tab:model_api_prices}
\end{table}

\subsection{Cost Metrics: Definitions and Implementation Details}
\label{app:cost_metrics}

This subsection supplements Sec.~\ref{sec:Experiments_and_Results} by providing
the formal definitions and implementation details of the cost-related metrics
reported for the model-first pipeline, including \emph{cost savings},
\emph{model contribution}, and the \emph{contribution ratio} (return per dollar
spent on model API usage and human rework).

Let
\begin{equation}
    B = \sum_{t \in \mathcal{T}} \mathrm{Price}(t),
\end{equation}
denote the human-only outsourcing cost over the evaluated split, where
$\mathcal{T}$ is the set of tasks and $\mathrm{Price}(t)$ is the contract price
of task $t$. For a model $m$ and settlement scenario $c$, let
$\mathrm{Success}_c(m,t) \in \{0,1\}$ indicate whether \emph{all} required
deliverables of task $t$ produced by model $m$ are accepted under scenario $c$
(i.e., the task is completed end-to-end without human redo). Let
$\mathrm{Cost}_{\mathrm{API}}(m)$ denote the total API spend of model $m$ on the
same split.

Under a model-first workflow, the expected total spend consists of the model API
cost plus human rework for failed tasks:
\[
\mathrm{Cost}_{\mathrm{API}}(m)
+ \sum_{t \in \mathcal{T}} \mathrm{Price}(t)\bigl(1 - \mathrm{Success}_c(m,t)\bigr).
\]
In our implementation, we approximate the per-call API prices using the assumptions in Appendix~\ref{subsection:API_Price_Assumptions}. We define the cost-savings ratio as
\begin{equation}
\resizebox{0.895\linewidth}{!}{$
    S_c(m)
    = 1
    - \frac{\mathrm{Cost}_{\mathrm{API}}(m)}{B}
    - \frac{1}{B} \sum_{t \in \mathcal{T}} \mathrm{Price}(t)\bigl(1 - \mathrm{Success}_c(m,t)\bigr),
$}
\end{equation}
which measures the overall outsourcing cost saved relative to the human-only
baseline.

The model contribution measures what fraction of the total contract value is
completed end-to-end by the model without human redo:
\begin{equation}
    M_c(m)
    = \frac{1}{B} \sum_{t \in \mathcal{T}} \mathrm{Price}(t)\,\mathrm{Success}_c(m,t).
\end{equation}

Finally, we report the contribution ratio, which captures the return per dollar
spent on model API usage and human rework under scenario $c$:
\begin{equation}
\resizebox{0.895\linewidth}{!}{$
    R_c(m)
    = \frac{\sum_{t \in \mathcal{T}} \mathrm{Price}(t)\,\mathrm{Success}_c(m,t)}
    {\mathrm{Cost}_{\mathrm{API}}(m) + \sum_{t \in \mathcal{T}} \mathrm{Price}(t)\bigl(1 - \mathrm{Success}_c(m,t)\bigr)}.
$}
\end{equation}

In the main text and Appendix Table~\ref{tab:cost_reduction}, we report $S_c(m)$
as \textit{Cost Savings (\%)}, $M_c(m)$ as Model Contribution (\%), and
\begin{table}[htbp]
  \centering
  \resizebox{\columnwidth}{!}{%
  \begin{tabular}{lrrr}
    \toprule
    \textbf{Model} & \textbf{\makecell{Cost \\ Savings \\ (\%)}} & \textbf{\makecell{Model \\ Contribution \\ (\%)}} & \textbf{\makecell{Contribution \\ Ratio}} \\
    \midrule

    \multicolumn{4}{c}{\textit{Closed-Source Models}} \\
    \midrule
    Gemini-Banana        &   37.0 &   37.0 & 0.5881 \\
    Gemini-Banana-Pro    & \cellcolor{rankFirst}  81.7 & \cellcolor{rankFirst}  81.7 & \cellcolor{rankFirst}4.4581 \\
    GPT-Image-1          &   50.2 &   50.3 & 1.0090 \\
    GPT-Image-1-Mini     &   38.3 &   38.3 & 0.6210 \\
    Ideogram-v3          &   27.3 &   27.3 & 0.3753 \\
    Imagen-4.0           &   37.8 &   37.8 & 0.6079 \\
    Kling-v2             &   32.6 &   32.6 & 0.4833 \\
    MJ-Relax-Edits       & \cellcolor{rankSecond}  58.9 & \cellcolor{rankSecond}  58.9 & \cellcolor{rankSecond}1.4349 \\
    MJ-Relax-Imagine     & \cellcolor{rankThird}  57.7 & \cellcolor{rankThird}  57.7 & \cellcolor{rankThird}1.3638 \\
    SeedEdit-3.0-i2i     &    0.6 &    0.6 & 0.0062 \\
    Seedream-3.0-t2i     &   35.6 &   35.7 & 0.5541 \\
    Seedream-4.0         &   22.3 &   22.4 & 0.2878 \\
    \midrule
    \multicolumn{4}{c}{\textbf{Open-Source Models}} \\
    \midrule
    FLUX-1.1-Pro         &   10.3 &   10.3 & 0.1154 \\
    FLUX-Kontext-Pro     &   18.0 &   18.0 & 0.2196 \\
    Qwen-Image           &   25.6 &   25.6 & 0.3438 \\
    SD-3.5-Large         &    5.7 &    5.8 & 0.0611 \\
    \bottomrule
  \end{tabular}%
  }
    \caption{Cost reduction under the standard settlement scenario.
  \textbf{Color:}
\legendbox{rankFirst}\,1st
\ \legendbox{rankSecond}\,2nd
\ \legendbox{rankThird}\,3rd.
  }
  \label{tab:cost_reduction}
\end{table}

\clearpage
\onecolumn         
\section{Task cases}
\label{app:task_cases}

\begin{center}
    \includegraphics[width=\textwidth]{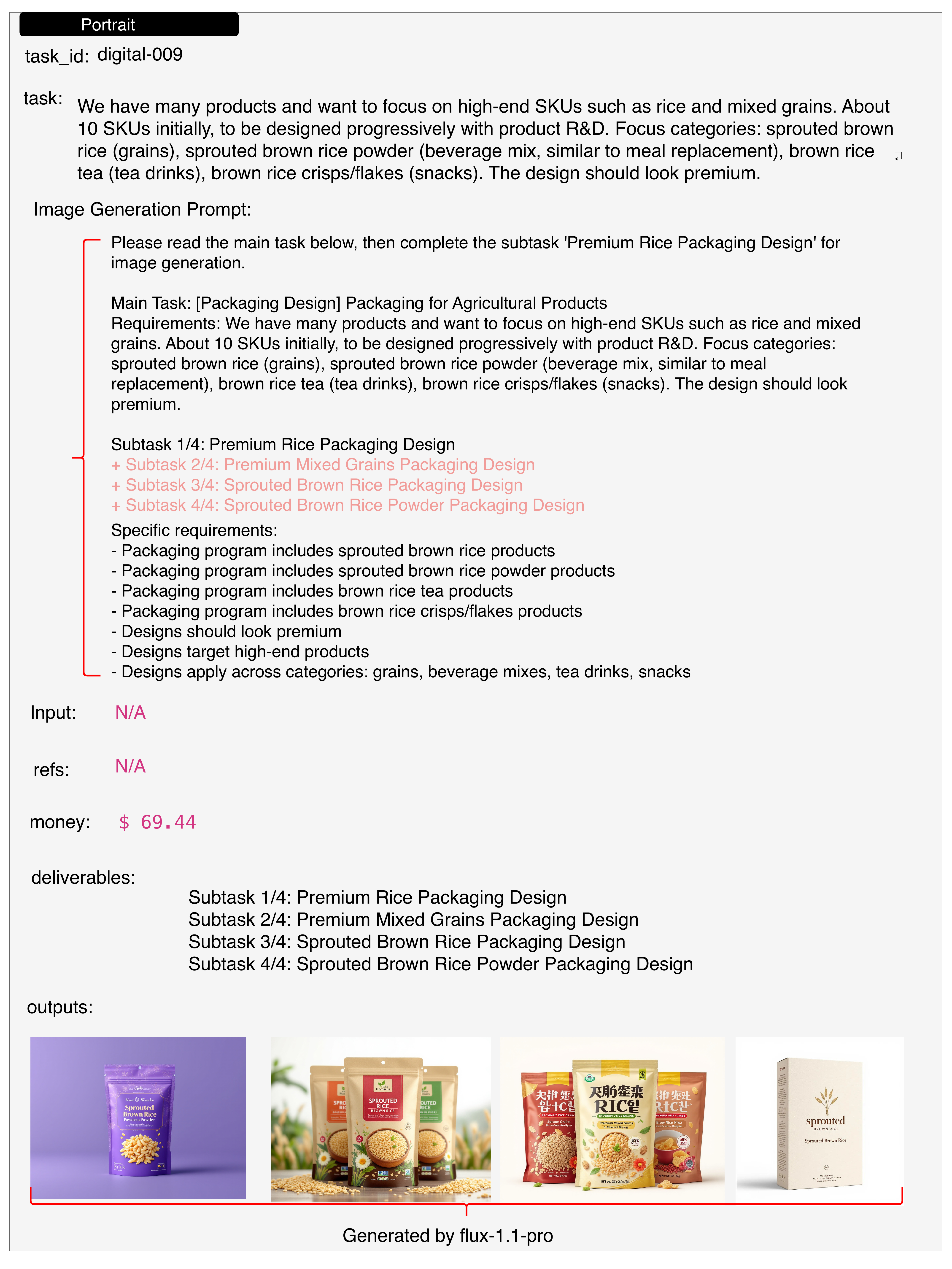}
    \captionof{figure}{Task case 1.}
    \label{fig:Task_case_1}
\end{center}

\clearpage

\begin{center}
    \includegraphics[width=\textwidth]{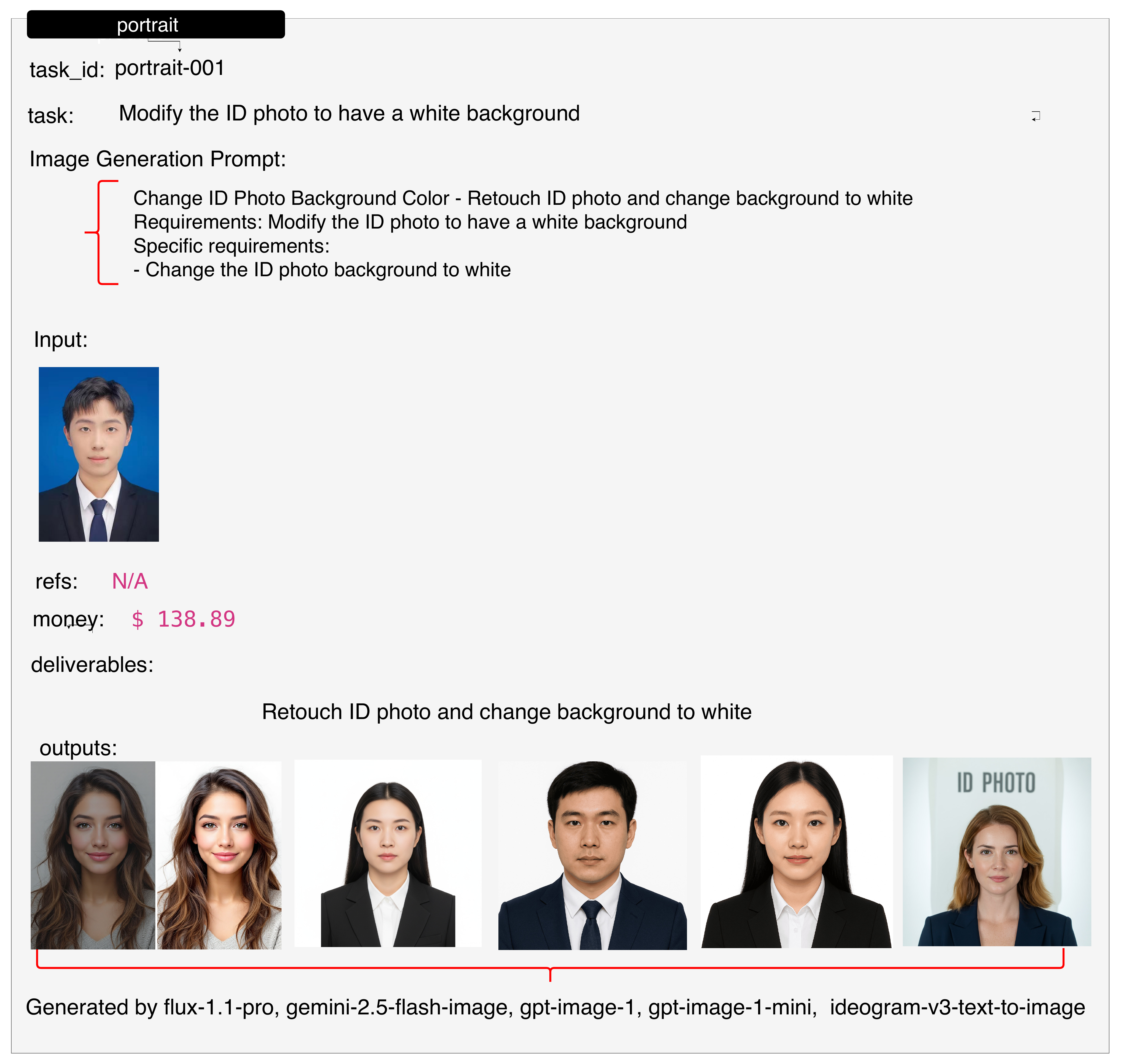}
    \captionof{figure}{Task case 2.}
    \label{fig:Task_case_2}
\end{center}

\clearpage
\begin{center}
    \includegraphics[width=\textwidth]{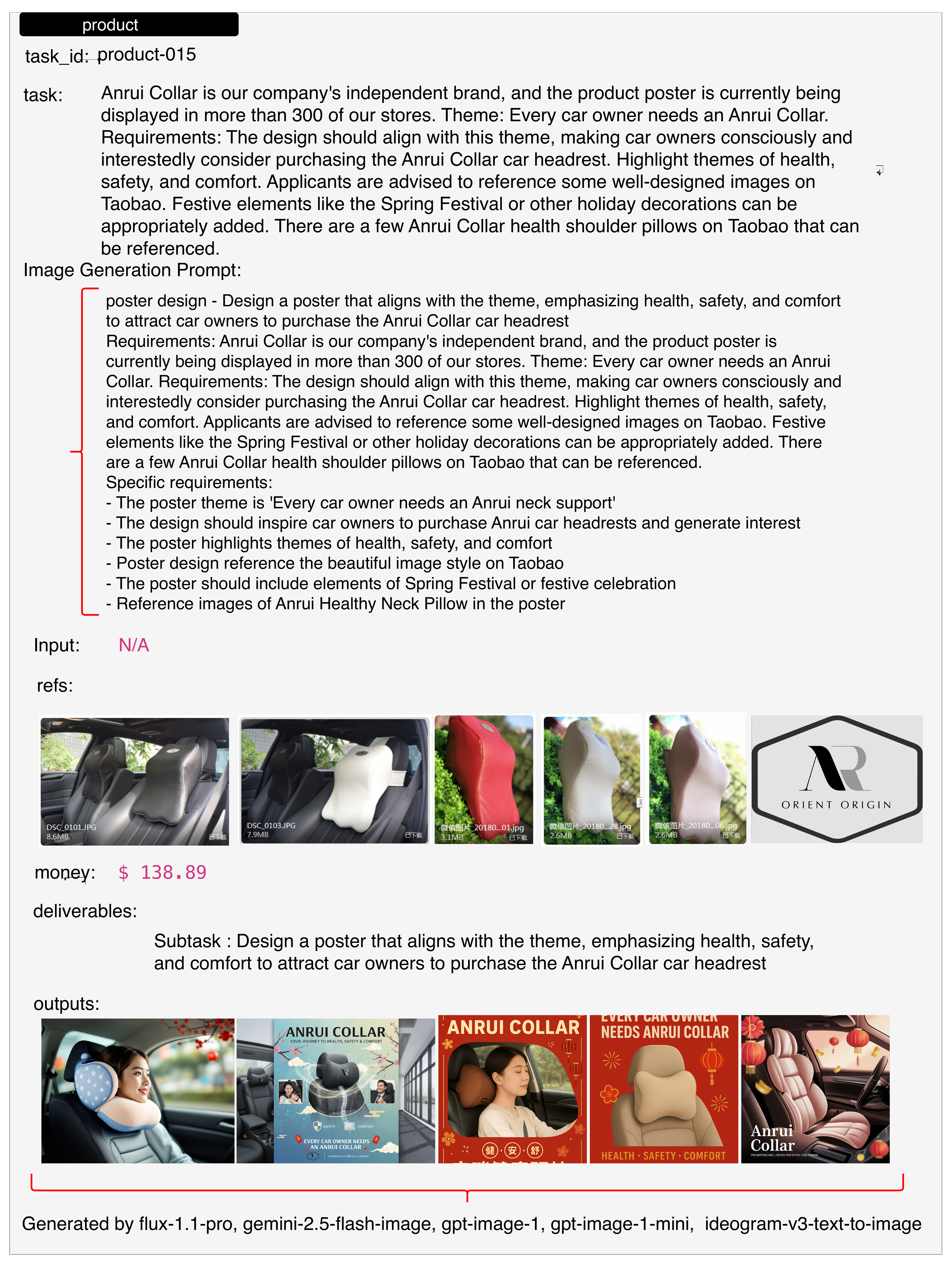}
    \captionof{figure}{Task case 3.}
    \label{fig:Task_case_3}
\end{center}

\clearpage
\twocolumn

\clearpage
\onecolumn
\section{More Result}

\begin{center}
    \footnotesize
    {\setlength{\tabcolsep}{10pt}
     \begin{tabular}{llcccc}
        \toprule
        \textbf{Model} & \textbf{Category} & \textbf{Rev (\$k)} & \textbf{Share (\%)} & \textbf{Task Acc. (\%)} & \textbf{Deliv. Acc. (\%)} \\
        \midrule
        \multicolumn{6}{c}{\emph{Closed-Source Models}} \\
        \midrule
        \multirow{3}{*}{Gemini-2.5-Flash-Image}
                                  & Portrait &   2.08 &  13.6 & 12.18 & 14.58 \\
                                  & Product  &  50.72 &  34.5 & 30.89 & 23.96 \\
                                  & Digital  &  56.48 &  42.5 & 42.41 & 50.49 \\
        \cmidrule{1-6}
        \multirow{3}{*}{Gemini-Banana-Pro}
                                  & Portrait &   3.90 &  25.5 & 25.46 & 25.00 \\
                                  & Product  & 122.60 &  83.4 & 85.81 & 89.17 \\
                                  & Digital  & 117.48 &  88.5 & 85.39 & 88.40 \\
        \cmidrule{1-6}
        \multirow{3}{*}{GPT-Image-1}
                                  & Portrait &   7.38 &  48.3 & 46.49 & 45.83 \\
                                  & Product  &  82.95 &  56.4 & 54.22 & 52.68 \\
                                  & Digital  &  57.99 &  43.7 & 41.83 & 40.81 \\
        \cmidrule{1-6}
        \multirow{3}{*}{GPT-Image-1-Mini}
                                  & Portrait &   0.17 &   1.1 &  1.48 &  1.39 \\
                                  & Product  &  87.95 &  59.8 & 57.11 & 62.81 \\
                                  & Digital  &  24.93 &  18.8 & 24.27 & 25.85 \\
        \cmidrule{1-6}
        \multirow{3}{*}{Ideogram-v3}
                                  & Portrait &   1.01 &   6.6 &  5.54 &  6.25 \\
                                  & Product  &  20.94 &  14.2 & 21.56 & 18.71 \\
                                  & Digital  &  58.57 &  44.1 & 47.28 & 51.47 \\
        \cmidrule{1-6}
        \multirow{3}{*}{Imagen-4.0}
                                  & Portrait &   4.03 &  26.3 & 26.94 & 27.08 \\
                                  & Product  &  70.46 &  47.9 & 34.44 & 30.87 \\
                                  & Digital  &  37.09 &  27.9 & 29.80 & 25.90 \\
        \cmidrule{1-6}
        \multirow{3}{*}{Kling-v2}
                                  & Portrait &   9.79 &  64.0 & 63.10 & 62.50 \\
                                  & Product  &  39.33 &  26.8 & 29.18 & 22.32 \\
                                  & Digital  &  47.02 &  35.4 & 36.68 & 24.27 \\
        \cmidrule{1-6}
        \multirow{3}{*}{MJ-Relax-Edits}
                                  & Portrait &   1.10 &   7.2 &  7.38 &  6.94 \\
                                  & Product  &  30.93 &  21.0 & 20.67 & 17.64 \\
                                  & Digital  &  46.78 &  35.2 & 36.39 & 39.39 \\
        \cmidrule{1-6}
        \multirow{3}{*}{MJ-Relax-Imagine}
                                  & Portrait &   8.82 &  57.7 & 61.25 & 57.64 \\
                                  & Product  &  48.20 &  32.8 & 29.78 & 25.03 \\
                                  & Digital  &  31.06 &  23.4 & 29.51 & 21.44 \\
        \cmidrule{1-6}
        \multirow{3}{*}{SeedEdit-3.0-i2i}
                                  & Portrait &   0.00 &   0.0 &  0.00 &  0.00 \\
                                  & Product  &   0.67 &   0.5 &  6.76 &  6.43 \\
                                  & Digital  &   1.14 &   0.9 &  7.41 &  7.27 \\
        \cmidrule{1-6}
        \multirow{3}{*}{Seedream-3.0-t2i}
                                  & Portrait &   0.95 &   6.2 &  5.54 &  5.90 \\
                                  & Product  &  71.46 &  48.6 & 48.89 & 55.07 \\
                                  & Digital  &  32.80 &  24.7 & 28.94 & 40.37 \\
        \cmidrule{1-6}
        \multirow{3}{*}{Seedream-4.0}
                                  & Portrait &   2.44 &  15.9 & 17.71 & 17.71 \\
                                  & Product  &  46.45 &  31.6 & 54.22 & 60.91 \\
                                  & Digital  &  17.06 &  12.8 & 11.75 & 28.29 \\
        \midrule
        \multicolumn{6}{c}{\textit{Open-Source Models}} \\
        \midrule
        \multirow{3}{*}{FLUX-1.1-Pro}
                                  & Portrait &   0.33 &   2.2 &  2.58 &  2.43 \\
                                  & Product  &   3.30 &   2.2 &  4.22 &  2.50 \\
                                  & Digital  &  26.91 &  20.3 & 23.21 & 16.65 \\
        \cmidrule{1-6}
        \multirow{3}{*}{FLUX-Kontext-Pro}
                                  & Portrait &   0.21 &   1.4 &  1.48 &  1.39 \\
                                  & Product  &  20.42 &  13.9 & 19.20 & 12.01 \\
                                  & Digital  &  32.51 &  24.5 & 25.00 & 20.00 \\
        \cmidrule{1-6}
        \multirow{3}{*}{Qwen-Image}
                                  & Portrait &   1.73 &  11.3 &  9.59 &  9.72 \\
                                  & Product  &  31.58 &  21.5 & 28.44 & 36.00 \\
                                  & Digital  &  42.19 &  31.8 & 32.38 & 31.77 \\
        \cmidrule{1-6}
        \multirow{3}{*}{SD-3.5-Large}
                                  & Portrait &   2.33 &  15.2 & 17.34 & 16.32 \\
                                  & Product  &   5.33 &   3.6 &  5.56 &  3.10 \\
                                  & Digital  &   9.33 &   7.0 &  5.73 &  6.31 \\
        \bottomrule
    \end{tabular}}
        \captionof{table}{Model performance on ServImage by category under the standard settlement scenario.
    \textbf{Revenue} is the total contract value earned by a model in each category.
    \textbf{Task Acceptance} and \textbf{Deliverable Acceptance} are the within-category proportions of tasks and deliverables approved according to human payment decisions ($y_{t,i}$).
    All monetary values are reported in thousand USD.}
    \label{tab:main_results_by_category}

\end{center}

\clearpage
\section{Prompting}
\label{app:prompting}
\subsection{Evaluation Points Execution Prompt}

\begin{center}
    \includegraphics[width=\textwidth]{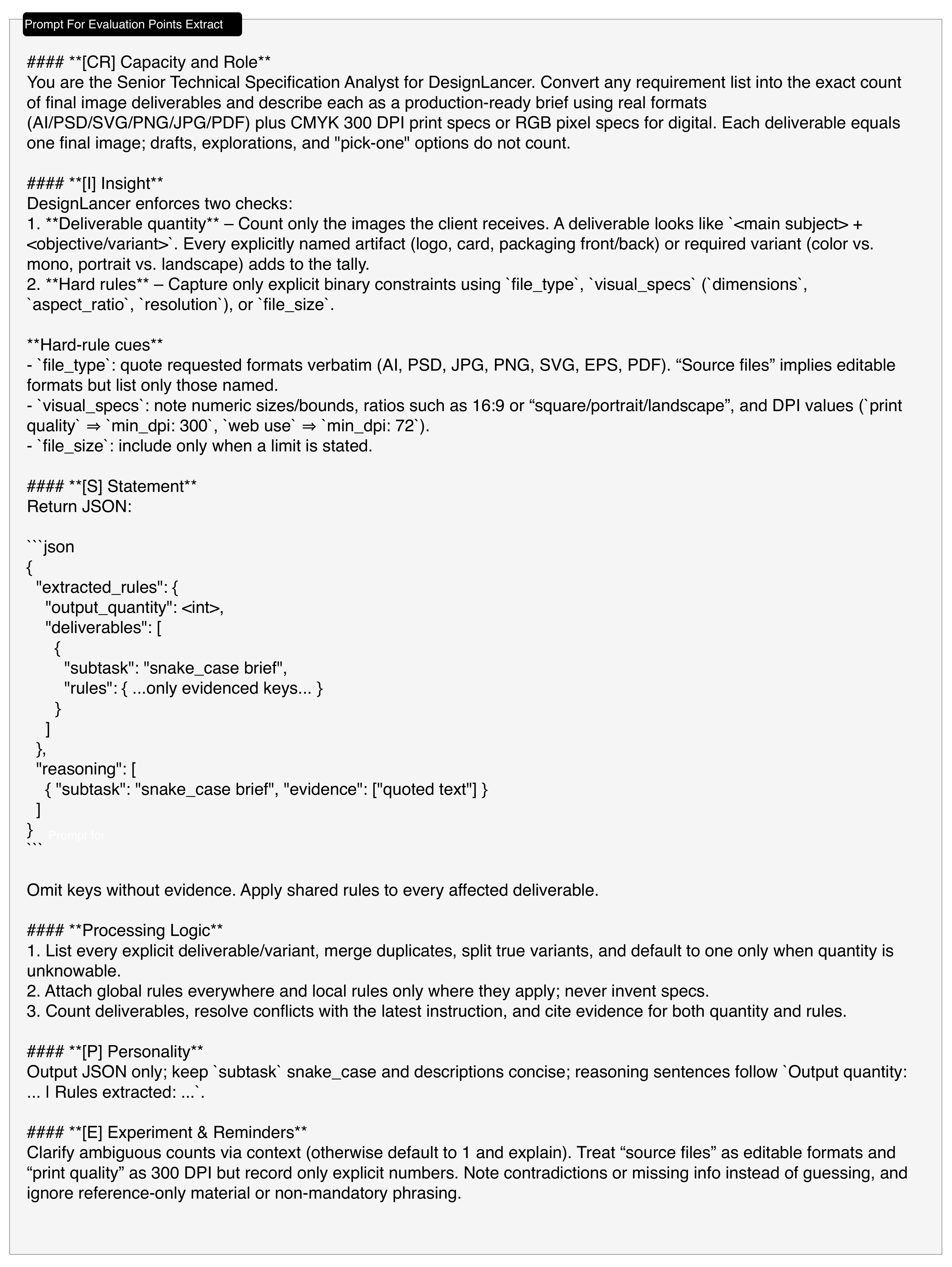}
    \captionof{figure}{Prompt for evaluation points extraction.}
    \label{fig:Prompt_For_Evaluation_Points_Extract}
\end{center}

\clearpage

\clearpage
\onecolumn          

\subsection{Evaluation Prompts}

\begin{center}
    \includegraphics[width=\textwidth]{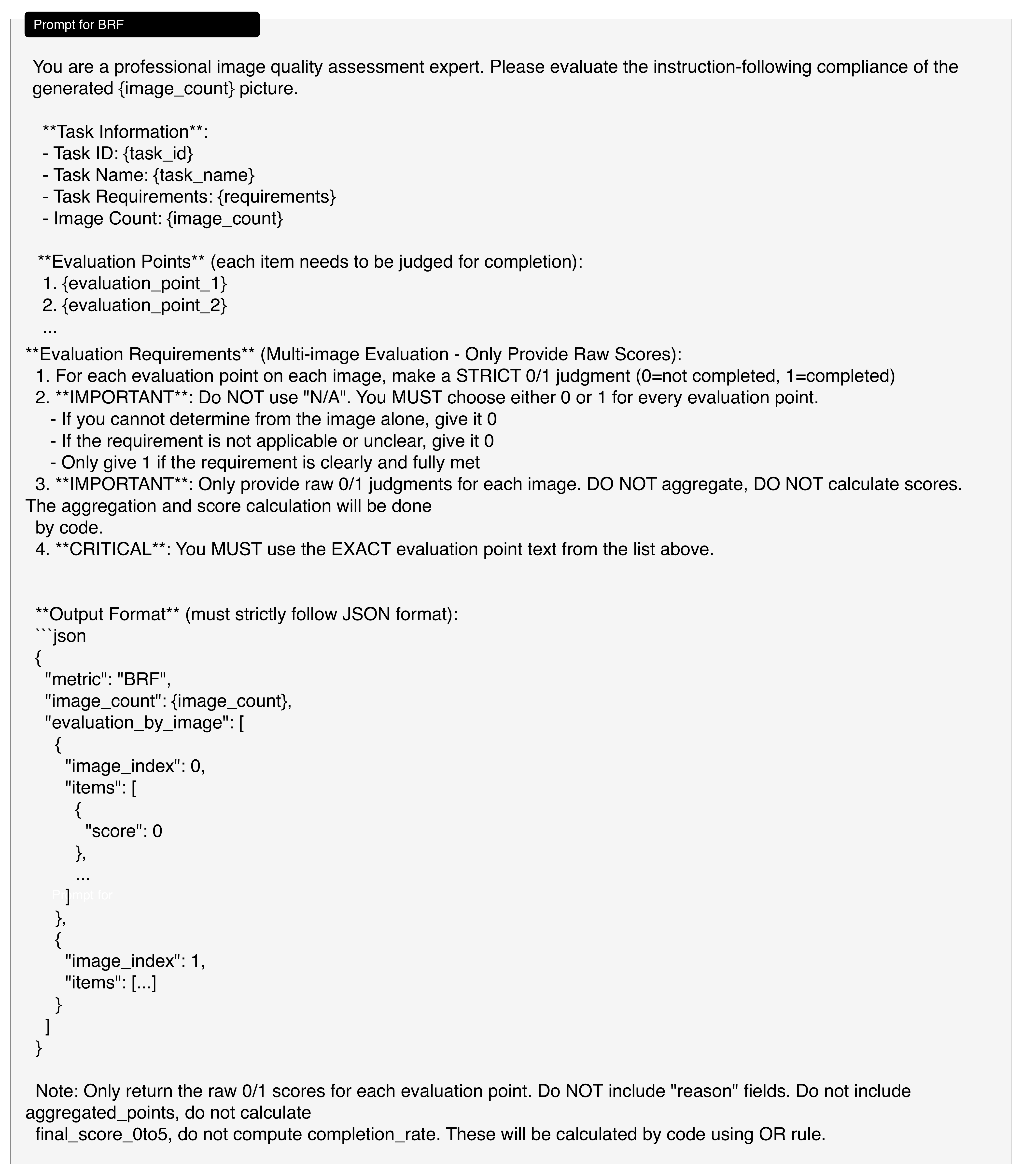}
    \captionof{figure}{BRF Evaluation prompt.}
    \label{fig:eval-prompt-brf}
\end{center}

\clearpage
\begin{center}
    \includegraphics[width=\textwidth]{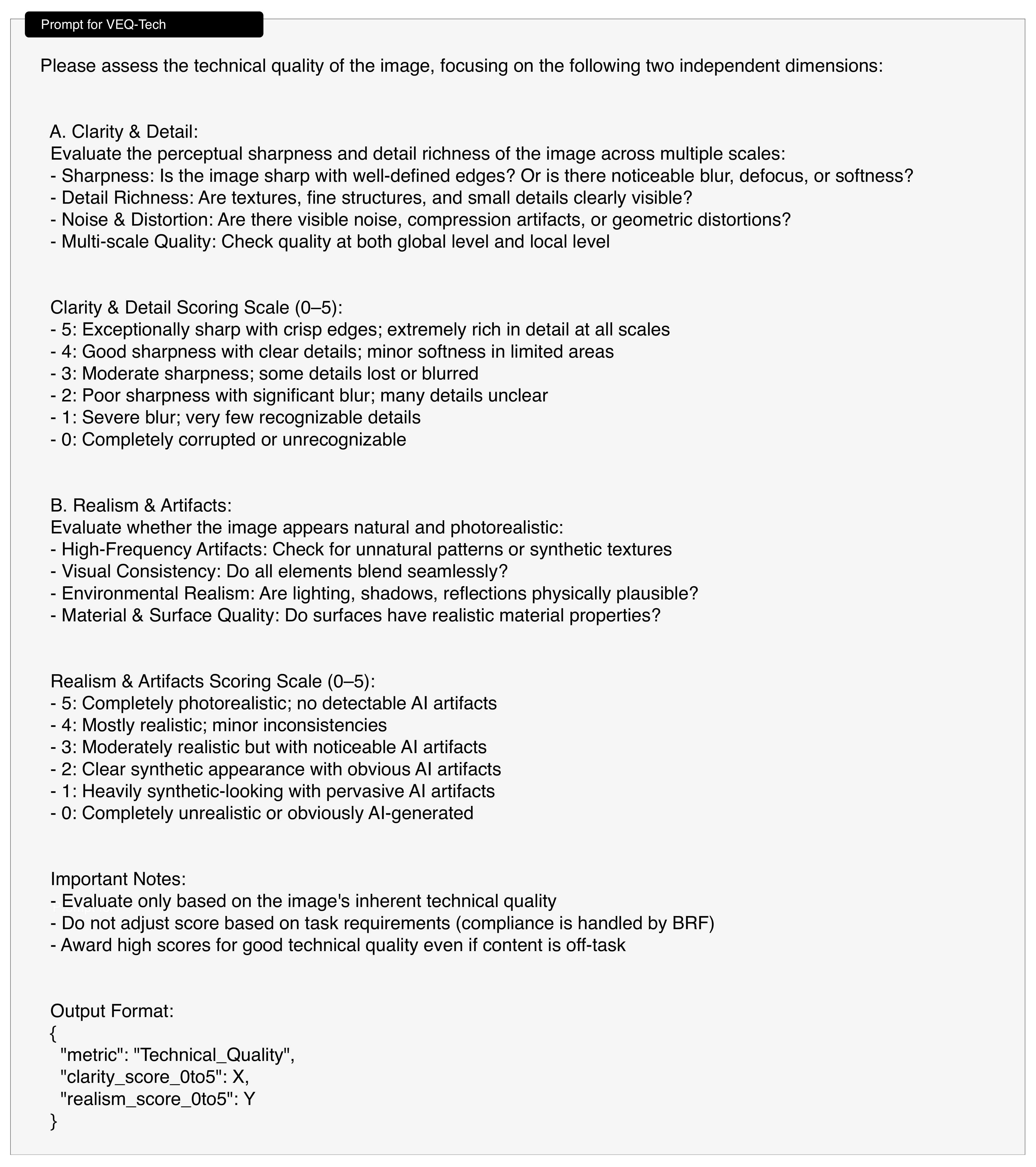}
    \captionof{figure}{VEQ-Tech Evaluation prompt.}
    \label{fig:eval-prompt-veq-tech}
\end{center}

\clearpage
\begin{center}
    \includegraphics[width=\textwidth]{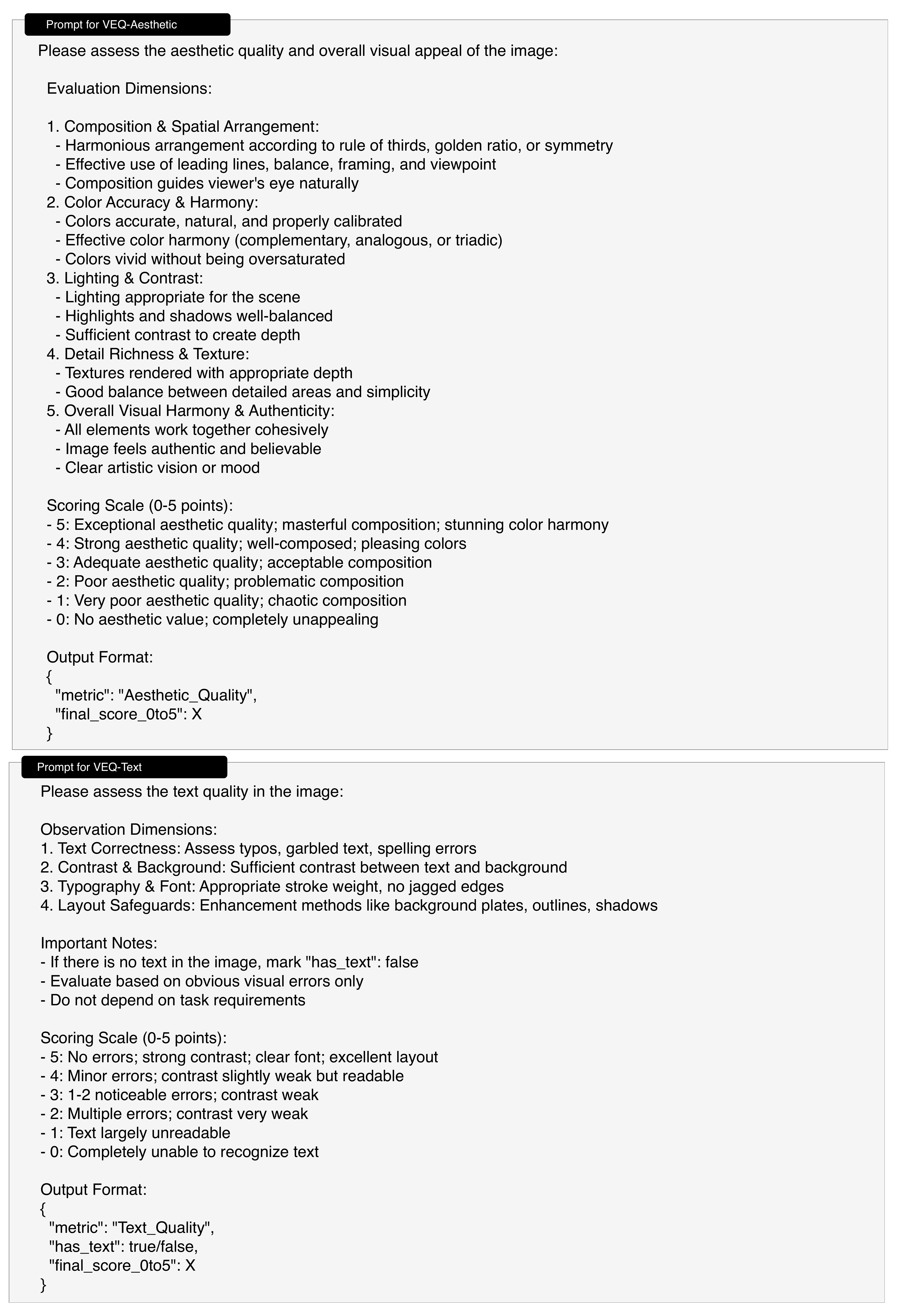}
    \captionof{figure}{VEQ-Aesthetic Quality AND Text Quality Evaluation prompt.}
    \label{fig:eval-prompt-veq}
\end{center}

\clearpage
\begin{center}
    \includegraphics[width=\textwidth]{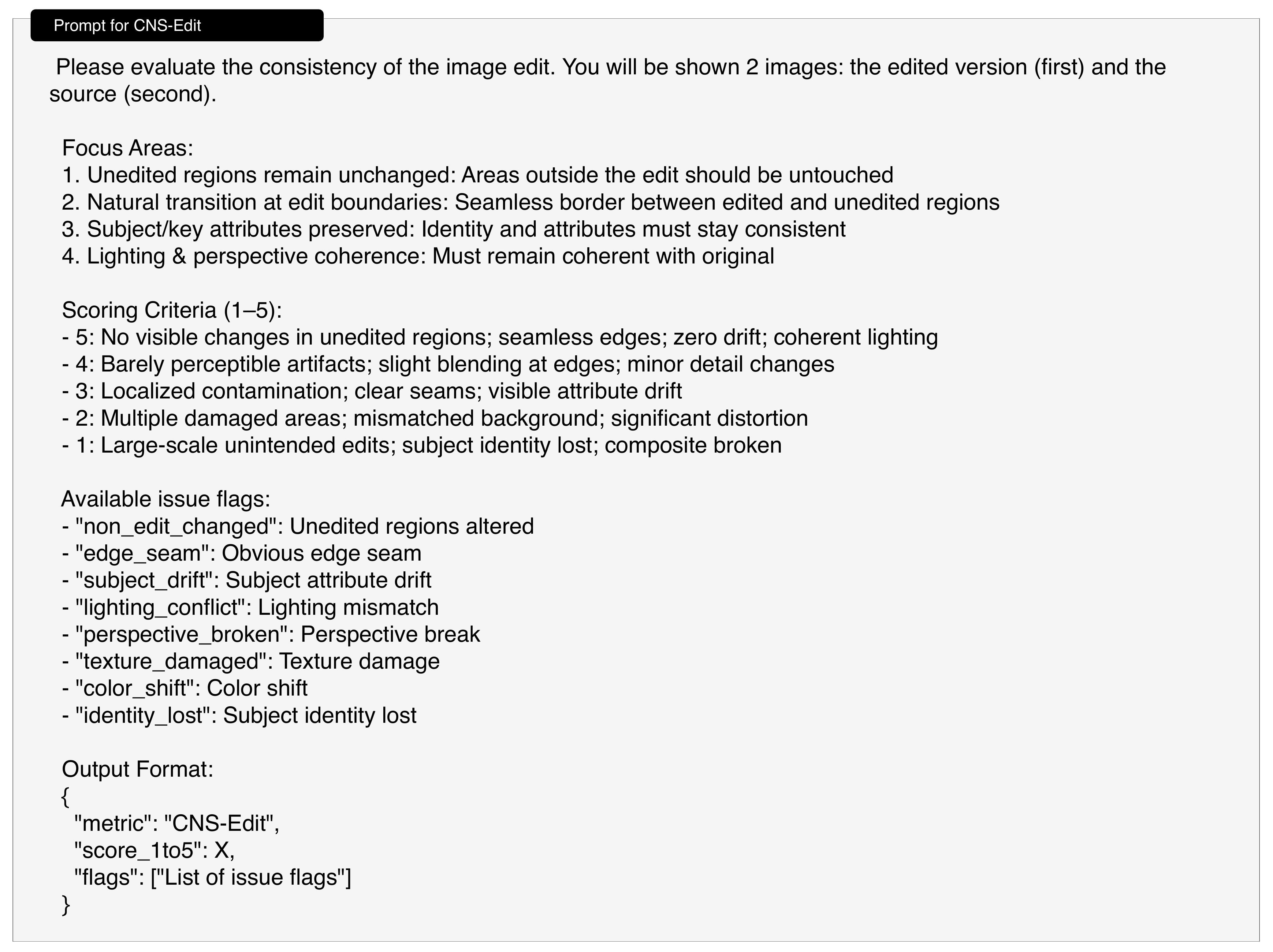}
    \captionof{figure}{CNS-Edit Evaluation prompt.}
    \label{fig:eval-prompt-CNS-Edit}
\end{center}

\clearpage
\begin{center}
    \includegraphics[width=\textwidth]{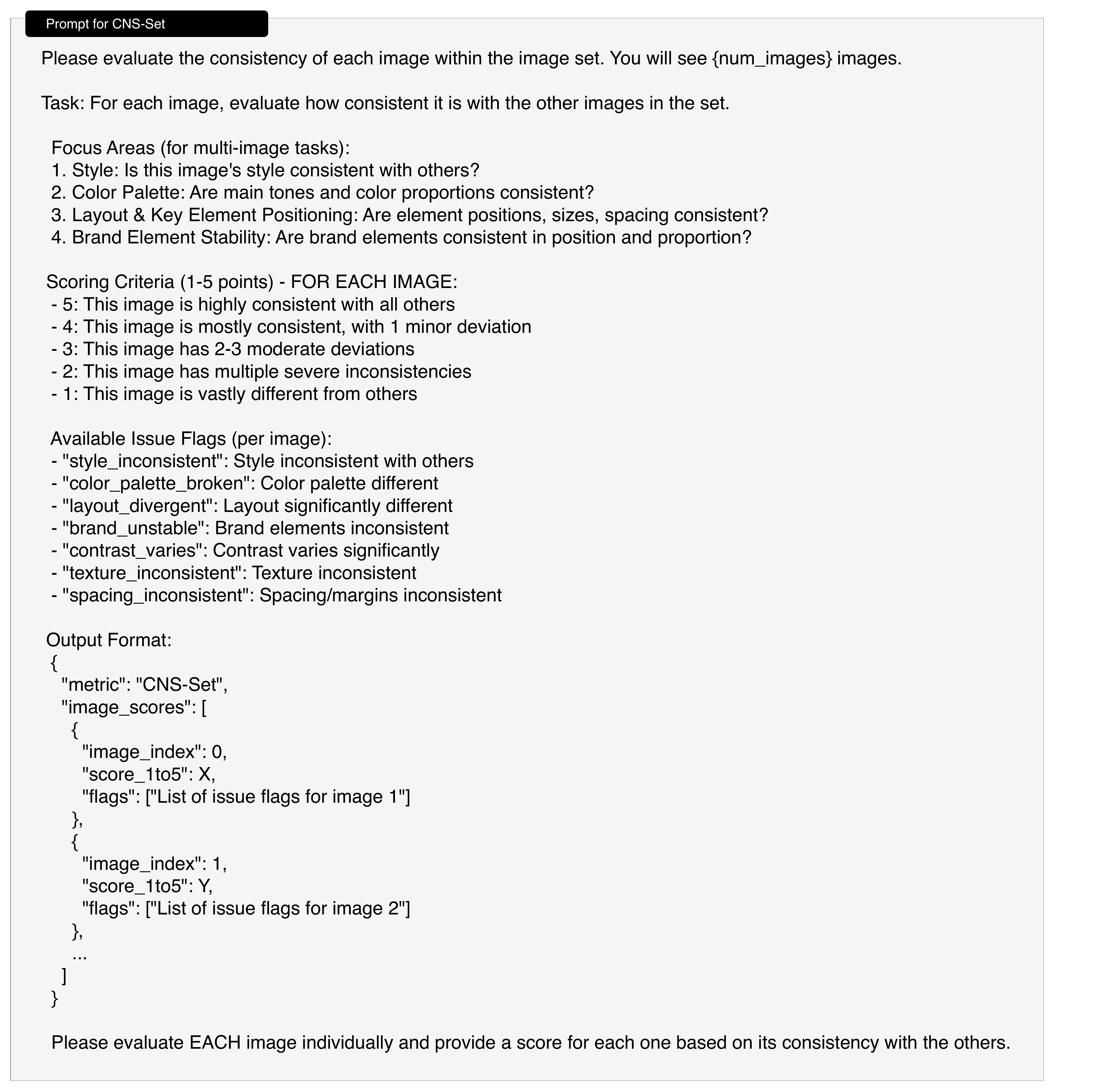}
    \captionof{figure}{CNS-Set Evaluation prompt.}
    \label{fig:eval-prompt-CNS-Set}
\end{center}

\twocolumn 

\section{Train Dataset}
We start from the original annotations: about 17k paid design tasks with roughly 32.9k accept/reject-labeled instances, and about 16k tasks with roughly 30.4k instances for concept supervision (BRF/VEQ/CNS). We remove ambiguous or low-confidence samples during preprocessing to reduce label noise. To mitigate accept/reject class imbalance, we apply training-only oversampling of accepted examples, while keeping the validation and test splits unchanged to reflect the natural accept/reject distribution.

\section{ServImageModel Settlement}
\label{sec:servimage_model_settlement}

\subsection{Training Setup}
\label{subsec:servimage_model_training_setup}
We adopt a two-stage training strategy on 4×A6000 GPUs to learn both image quality assessment and downstream accept/reject decision-making.
In Stage 1, on the same 4×A6000 GPUs, we fine-tune the model for 5 epochs to predict seven-dimensional quality scores using AdamW (lr $4\times10^{-5}$, weight decay 0.01, cosine schedule with 3\% warmup), with an effective batch size of 32 (batch size 1, accumulation 8).
LoRA is applied to the vision encoder, projector, and language model (rank $r{=}16$, $\alpha{=}16$, dropout 0.01), and we compress auxiliary images into collages to control visual tokens (max sequence length 8,192). Training uses DeepSpeed with BF16 mixed precision and gradient checkpointing, saving and evaluating every 500 steps. 
In Stage 2, on the same 4×A6000 GPUs, we freeze Stage 1 weights and train a decision module for 2 epochs with a larger effective batch size of 64 (per-device batch size 2, accumulation 8) and a learning rate $4\times10^{-5}$, adding lightweight LoRA adapters (dropout 0.01).
The decision head fuses representations from the frozen quality model and the original VLM by concatenation for binary classification. 

\subsection{Automatic Settlement Results}
\label{subsec:servimage_model_settlement_results}
ServImageModel-based settlement is an automatic proxy. It converts predicted payment probabilities into accept/reject outcomes and then applies the same accounting rule as in Table~\ref{tab:main_results}. Because this pipeline involves thresholding and task-level aggregation, its absolute revenues can deviate from human settlement, especially on particular categories; therefore, we use it only for scalable approximation, while all main conclusions rely on human labels.

\begin{table*}[!t]
    \centering
    \scriptsize
    \setlength{\tabcolsep}{2pt}
    \renewcommand{\arraystretch}{0.85}
    \begin{tabular}{lrrrrrrrrrr}
        \toprule
        \multirow{2}{*}{\textbf{Model}} &
        \multicolumn{4}{c}{\textbf{Total}} &
        \multicolumn{2}{c}{\textbf{Portrait}} &
        \multicolumn{2}{c}{\textbf{Product}} &
        \multicolumn{2}{c}{\textbf{Digital}} \\
        \cmidrule(lr){2-5}\cmidrule(lr){6-7}\cmidrule(lr){8-9}\cmidrule(lr){10-11}
        & \textbf{Revenue (\$k)} & \textbf{Share (\%)} &
          \textbf{Task Acc. (\%)} & \textbf{Deliv. Acc. (\%)} &
          \textbf{Rev (\$k)} & \textbf{Share (\%)} &
          \textbf{Rev (\$k)} & \textbf{Share (\%)} &
          \textbf{Rev (\$k)} & \textbf{Share (\%)} \\
        \midrule
        \multicolumn{11}{c}{\emph{Closed-Source Models}} \\
        \midrule
        Gemini-Banana      & 149.82 & 50.8 & 53.76 & 64.38 & 8.91 & 58.3 & 84.69 & 57.6 & 56.22 & 42.3 \\
        Gemini-Banana-Pro  & \cellcolor{rankFirst}199.24 & \cellcolor{rankFirst}67.5 & \cellcolor{rankFirst}73.85 & \cellcolor{rankSecond}74.96 & 0.03 & 0.2 & \cellcolor{rankFirst}143.99 & \cellcolor{rankFirst}98.0 & 55.21 & 41.6 \\
        GPT-Image-1        & 145.24 & 49.2 & 55.34 & 65.23 & \cellcolor{rankThird}10.32 & \cellcolor{rankThird}67.5 & 78.88 & 53.7 & 56.04 & 42.2 \\
        GPT-Image-1-Mini   & 147.93 & 50.1 & 56.25 & 65.30 & 9.62 & 63.0 & 83.10 & 56.5 & 55.20 & 41.6 \\
        Ideogram-v3        & 158.02 & 53.6 & 56.09 & 65.18 & \cellcolor{rankSecond}10.37 & \cellcolor{rankSecond}67.9 & \cellcolor{rankThird}87.73 & \cellcolor{rankThird}59.7 & 59.91 & 45.1 \\
        Imagen-4.0         & \cellcolor{rankSecond}183.27 & \cellcolor{rankSecond}62.1 & \cellcolor{rankSecond}70.19 & \cellcolor{rankFirst}80.45 & \cellcolor{rankFirst}13.71 & \cellcolor{rankFirst}89.7 & 62.84 & 42.8 & \cellcolor{rankFirst}106.72 & \cellcolor{rankFirst}80.4 \\
        Kling-v2           & 29.64 & 10.0 & 19.55 & 24.40 & 7.13 & 46.6 & 1.82 & 1.2 & 20.69 & 15.6 \\
        MJ-Relax-Edits     & 139.95 & 47.4 & 54.44 & 65.38 & 9.40 & 61.5 & 70.22 & 47.8 & 60.33 & 45.4 \\
        MJ-Relax-Imagine   & 136.33 & 46.2 & 55.58 & 66.11 & 10.01 & 65.5 & 69.61 & 47.4 & 56.71 & 42.7 \\
        SeedEdit-3.0-i2i   & 18.37 & 6.2 & 50.39 & \cellcolor{rankThird}66.53 & 0.05 & 0.3 & 6.18 & 4.2 & 12.13 & 9.1 \\
        Seedream-3.0-t2i   & 144.60 & 49.0 & 55.07 & 66.19 & 10.23 & 66.9 & 73.99 & 50.3 & 60.38 & 45.5 \\
        Seedream-4.0       & 0.00 & 0.0 & 0.00 & 0.00 & 0.00 & 0.0 & 0.00 & 0.0 & 0.00 & 0.0 \\
        \midrule
        \multicolumn{11}{c}{\emph{Open-Source Models}} \\
        \midrule
        FLUX-1.1-Pro       & \cellcolor{rankThird}169.68 & \cellcolor{rankThird}57.5 & \cellcolor{rankThird}58.16 & 66.32 & 9.79 & 64.1 & \cellcolor{rankSecond}89.50 & \cellcolor{rankSecond}60.9 & \cellcolor{rankSecond}70.39 & \cellcolor{rankSecond}53.0 \\
        FLUX-Kontext-Pro   & 138.07 & 46.8 & 54.90 & 65.49 & 9.44 & 61.8 & 69.94 & 47.6 & 58.69 & 44.2 \\
        Qwen-Image         & 148.09 & 50.2 & 55.50 & 65.68 & 10.22 & 66.8 & 82.41 & 56.1 & 55.47 & 41.8 \\
        SD-3.5-Large       & 132.88 & 45.0 & 53.94 & 64.67 & 9.58 & 62.6 & 61.12 & 41.6 & \cellcolor{rankThird}62.18 & \cellcolor{rankThird}46.8 \\
        \bottomrule
    \end{tabular}
    \caption{Model performance on ServImage under the standard settlement scenario, \textbf{estimated using ServImageModel-predicted payment probabilities}.
Revenue, Rev (\$k), is the total contract value estimated from \emph{predicted} accept/reject outcomes, and Share is its fraction of the overall contract value (\$295k), while category shares are computed within each category.
Task Acceptance and Deliverable Acceptance are the predicted proportions of tasks and deliverables approved under ServImageModel.
(See Table~\ref{tab:main_results} for results based on \textbf{human} payment-decision labels.)
\textbf{Color:}
\legendbox{rankFirst}\,1st
\ \legendbox{rankSecond}\,2nd
\ \legendbox{rankThird}\,3rd.
}
    \label{tab:main_results_auto_by_serviamgemodel}
\end{table*}

\section{Judge Robustness and Reproducibility}
\label{app:Judge_Robustness_and_Reproducibility}
To further reduce reliance on a single fixed VLM judge, we conduct two additional analyses.
First, we perform human verification on a randomly sampled set of 300 deliverables and report the agreement between each automatic judge and human judgments on BRF, VEQ, CNS, and the overall score.
Second, we recompute ServImageScore using two additional judge models and compare the resulting system rankings across judges. We report the rank correlation and Top-$k$ overlap between judge pairs, showing that our main conclusions remain stable under different judge choices.

\begin{table}[t]
\centering
\small
\begin{tabular}{lcc}
\toprule
\textbf{Judge} & \textbf{BRF vs Human} & \textbf{VEQ vs Human} \\
               & \textbf{CNS vs Human} & \textbf{Overall vs Human} \\
\midrule
GPT-5-mini     & 0.87 \quad 0.88 & 0.87 \quad 0.88 \\
GPT-5          & 0.91 \quad 0.90 & 0.91 \quad 0.91 \\
Gemini-3-Flash & 0.80 \quad 0.89 & 0.84 \quad 0.85 \\
\bottomrule
\end{tabular}
\caption{
Agreement between automatic judges and human verification on 300 randomly sampled deliverables.
For each dimension (BRF, VEQ, and CNS), we compute the agreement rate as the proportion of instances for which the judge's decision matches the human judgment.
\textbf{Overall vs Human} is computed in the same way after aggregating the three dimensions into the final overall decision for each deliverable.
}
\label{tab:judge_human_agreement}
\end{table}

\begin{table}[t]
\centering
\small
\resizebox{\columnwidth}{!}{%
\begin{tabular}{lcccc}
\toprule
\textbf{Pair of Judges} & $\rho_{\mathrm{BRF}}$ & $\rho_{\mathrm{VEQ}}$ & $\rho_{\mathrm{CNS}}$ & $\rho_{\mathrm{overall}}$ \\
\midrule
GPT-5-mini vs GPT-5           & 0.86 & 0.74 & 0.82 & 0.87 \\
GPT-5-mini vs Gemini-3-Flash  & 0.77 & 0.69 & 0.77 & 0.79 \\
GPT-5 vs Gemini-3-Flash       & 0.76 & 0.70 & 0.77 & 0.80 \\
\bottomrule
\end{tabular}%
}
\caption{
Rank correlation between judge pairs over model rankings induced by each judge.
For each dimension, $\rho$ denotes the Spearman rank correlation coefficient computed from the full ranking of models under the corresponding judge.
The overall ranking is obtained by recomputing ServImageScore with each judge and then measuring the Spearman correlation between the resulting overall rankings.
In addition, we also examine Top-$k$ overlap, defined as the proportion of shared models in the top-$k$ positions between two judge-specific rankings.
}
\label{tab:judge_rank_correlation}
\end{table}

\section{Failure Analyze}
\label{app:Failure_Analyze}
In this appendix, we provide a more systematic analysis of rejection causes and the payment boundary, addressing the concern that the original case study was overly qualitative and insufficiently clear about why samples were rejected. Specifically, for all rejected samples ($N = 21{,}557$), we assign each case to its lowest-scoring dimension, which we treat as the dominant failure dimension. We then report, for each dimension, its frequency among rejected samples, its category-wise distribution, and the mean scores for accepted and rejected samples, together with Cohen's $d$ as an effect-size measure. The results show that BRF is the most frequent failure mode (53.8\%) and also the dimension that best separates accepted from rejected samples ($d = 0.39$), suggesting that the main bottleneck lies in specification and constraint satisfaction. We also observe distinct category-level patterns: Portrait failures are more often associated with \textsc{VEQ-Realism} (23.7\%) and \textsc{VEQ-Clarity} (17.7\%), Product failures are dominated by BRF (64.7\%), and Digital/Art failures are mainly driven by BRF (48.3\%) and CNS (30.3\%), highlighting the importance of cross-image consistency in such tasks. By contrast, \textsc{VEQ-Aesthetic} barely separates accepted and rejected outcomes ($d=-0.02$), further suggesting that requirement satisfaction and consistency, rather than aesthetic preference alone, more strongly define the payment boundary.

To further illustrate these failure modes, we provide representative failure examples in Table~\ref{tab:failure_examples}. These examples show that rejection may arise from missing required elements (BRF), insufficient clarity or realism, severe aesthetic degradation, text rendering errors, or poor consistency across multiple generated images.

\begin{table*}[t]
\centering
\small
\resizebox{\textwidth}{!}{%
\begin{tabular}{lccccccc}
\toprule
\textbf{Dimension} & \textbf{Overall (\%)} & \textbf{Portrait (\%)} & \textbf{Product (\%)} & \textbf{Digital (\%)} & \textbf{Acc. Score} & \textbf{Rej. Score} & \textbf{Cohen's $d$} \\
\midrule
BRF              & 53.8 & 40.8 & 64.7 & 48.3 & 3.48 & 2.82 & 0.39 \\
VEQ: Clarity     & 5.1  & 17.7 & 2.4  & 3.0  & 4.69 & 4.54 & 0.26 \\
VEQ: Realism     & 9.6  & 23.7 & 4.1  & 9.7  & 4.29 & 4.22 & 0.11 \\
VEQ: Aesthetic   & 4.2  & 7.0  & 4.6  & 2.9  & 4.14 & 4.15 & -0.02 \\
VEQ: Text        & 5.2  & 5.5  & 4.6  & 5.8  & 4.33 & 4.17 & 0.18 \\
CNS              & 22.1 & 5.4  & 19.6 & 30.3 & 3.66 & 3.42 & 0.22 \\
\bottomrule
\end{tabular}%
}
\caption{
Systematic failure analysis over rejected samples ($N = 21{,}557$).
For each rejected sample, the \textbf{dominant failure dimension} is defined as the lowest-scoring dimension for that sample.
\textbf{Overall (\%)} denotes the proportion of all rejected samples whose dominant failure falls into the corresponding dimension.
\textbf{Portrait (\%)}, \textbf{Product (\%)}, and \textbf{Digital (\%)} denote the proportion of rejected samples within each category whose dominant failure is the corresponding dimension.
\textbf{Acc. Score} and \textbf{Rej. Score} are the mean scores of accepted and rejected samples, respectively, on that dimension.
\textbf{Cohen's $d$} is computed as the standardized mean difference between accepted and rejected samples on the corresponding dimension, i.e., the difference in means divided by the pooled standard deviation.
}
\label{tab:systematic_failure_analysis}
\end{table*}

\begin{table*}[t]
\centering
\small
\setlength{\tabcolsep}{4pt}
\renewcommand{\arraystretch}{1.05}
\begin{tabularx}{\textwidth}{l l l >{\raggedright\arraybackslash}X c >{\raggedright\arraybackslash}X}
\toprule
\textbf{Dimension} & \textbf{Model} & \textbf{Category} & \textbf{Task Description} & \textbf{Dominant Score} & \textbf{Failure Notes} \\
\midrule
BRF & kling-v2-images & Digital & Design 8 anime characters + expressions & 2.0 &
Missing some of the 8 characters; missing four expressions; missing weekday-to-character mapping \\
VEQ: Clarity & qwen-text-to-image & Portrait & Photo with professional gray-toned outfit & 4.0 &
BRF is perfect, but clarity, realism, and aesthetics are all 4 (tie $\rightarrow$ lowest dimension assignment) \\
VEQ: Realism & flux-1.1-pro & Digital & Bookworm animal reading scene & 4.0 &
BRF (=4.5), Clarity (=5), Aesthetic (=5), but Realism (=4) \\
VEQ: Aesthetic & mj-relax-imagine\_4 & Digital & Historical characters from Tang/Song/Ming & 0.0 &
Clarity (=4), Realism (=4), but Aesthetic (=0), indicating an extreme aesthetic failure \\
VEQ: Text & seedream-4.0 & Digital & Animal cartoon brand mascot & 2.0 &
Other dimensions are 4--5, but Text (=2), indicating a text rendering failure \\
CNS & flux-kontext-pro & Digital & Virtual fitness-coach IP design & 2.5 &
BRF (=4.29), Clarity (=5), but CNS (=2.5), indicating poor consistency across 12 images \\
\bottomrule
\end{tabularx}
\caption{
Representative failure examples grouped by dominant failure dimension.
\textbf{Dominant Score} denotes the score of the lowest-scoring dimension for the corresponding sample, which determines the failure label used in Table~\ref{tab:systematic_failure_analysis}.
When multiple dimensions tie for the minimum score, the example is assigned according to the predefined tie-breaking rule used in our analysis pipeline.
}
\label{tab:failure_examples}
\end{table*}

\end{document}